\documentclass[dvipsnames]{article} %
\usepackage{colm2024_conference}

\usepackage{booktabs}
\usepackage{enumitem}
\usepackage{wrapfig}
\usepackage{algorithm}
\usepackage{algpseudocode}
\usepackage{graphicx}
\usepackage{subfigure}
\usepackage[misc]{ifsym}
\usepackage{hyperref}

\usepackage{microtype}
\usepackage{amsmath}
\usepackage{colortbl}
\usepackage[utf8]{inputenc}
\usepackage[T1]{fontenc}
\definecolor{lightgray}{rgb}{0.9,0.9,0.9}
\usepackage{caption}
\usepackage{subcaption}
\usepackage{setspace}
\usepackage{url}
\usepackage{multirow}
\usepackage{tabularx}
\usepackage{blindtext}
\usepackage{pgfplots}
\pgfplotsset{compat=1.18} 
\usepackage{tikz}
\usetikzlibrary{er,positioning,bayesnet}
\usepackage{makecell}
\usepackage{tipa}
\usepackage{siunitx}
\usepackage{nicefrac}
\usepackage{listings}
\usepackage[raster,skins, most]{tcolorbox} %
\usepackage{xltabular}
\usepackage{adjustbox}
\usepackage{xurl}
\usepackage{rotating}
\usepackage{xcolor}
\usepackage[normalem]{ulem}
\usepackage[dvipsnames]{xcolor}

\usepackage{fontawesome}

\useunder{\uline}{\ul}{}

\usepackage{amsmath,amsfonts,bm}

\def\eqref#1{equation~\ref{#1}}

\def\1{\bm{1}}

\DeclareMathAlphabet{\mathsfit}{\encodingdefault}{\sfdefault}{m}{sl}
\SetMathAlphabet{\mathsfit}{bold}{\encodingdefault}{\sfdefault}{bx}{n}

\renewcommand{\ghlink}{https://github.com/Alibaba-NLP/DeepResearch}

\newcommand*\justify{%
  \fontdimen2\font=0.4em
  \fontdimen3\font=0.2em
  \fontdimen4\font=0.1em
  \fontdimen7\font=0.1em
  \hyphenchar\font=`\-
}

\renewcommand{\texttt}[1]{%
  \begingroup
  \ttfamily
  \begingroup\lccode`~=`/\lowercase{\endgroup\def~}{/\discretionary{}{}{}}%
  \begingroup\lccode`~=`[\lowercase{\endgroup\def~}{[\discretionary{}{}{}}%
  \begingroup\lccode`~=`.\lowercase{\endgroup\def~}{.\discretionary{}{}{}}%
  \catcode`/=\active\catcode`[=\active\catcode`.=\active
  \justify\scantokens{#1\noexpand}%
  \endgroup
}

\usepackage{makecell}
\usetikzlibrary{tikzmark}

\makeatletter
\newcommand*\myfontsize{%
  \@setfontsize\myfontsize{7}{8}%
}
\makeatother

\definecolor{uclablue}{RGB}{159, 195, 224}

\definecolor{uclagold}{RGB}{255, 240, 180}

\definecolor{aliceblue}{RGB}{255, 238, 241}

\definecolor{cadmiumgreen}{rgb}{0.0, 0.42, 0.24}

\definecolor{myred}{rgb}{0.7, 0.3, 0.0}
\definecolor{myblue}{rgb}{0.2, 0.3, 0.6}
\definecolor{babygreen}{rgb}{0.85, 0.97, 0.85}

\definecolor{purple1}{RGB}{126, 107, 196}
\definecolor{purple2}{RGB}{199, 158, 207}
\definecolor{purple3}{RGB}{214, 200, 255}
\definecolor{purple4}{RGB}{254, 240, 255}

\definecolor{deepblue}{RGB}{48, 58, 82}

\newcommand{\symboletongyi}{\raisebox{0pt}{~\includegraphics[scale=0.012]{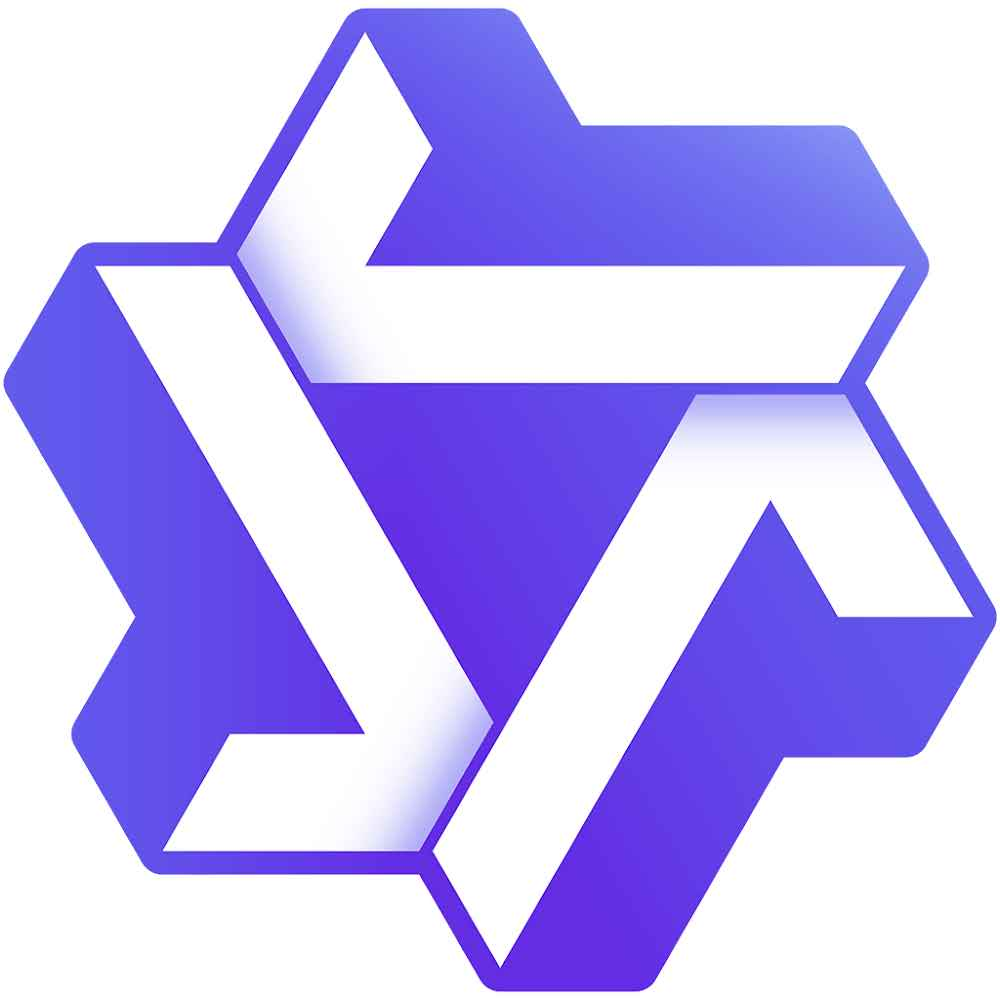}}~}

\definecolor{deepPurple}{HTML}{330066}

\definecolor{uclablue_old}{rgb}{0.15, 0.45, 0.68}
\hypersetup{
    breaklinks,
    citecolor=uclablue_old,
    colorlinks=true,
}

\newtcolorbox{mybox}[2][]
  {colback = black!5!white, colframe = black!75!black, fonttitle = \bfseries,
    colbacktitle = black!100!black, enhanced, before upper={\fontsize{8}{11}\obeyspaces\obeylines\selectfont}, fontupper=\selectfont,
    attach boxed title to top left={yshift=-2.2mm,xshift=4mm},
    title=#2,#1}

\definecolor{mydarkpurple}{RGB}{102,50,227}   
\definecolor{mylightpurple}{RGB}{230, 230, 250} 

\title{%
\raisebox{-1.2em}{
  \hspace*{-0.2in} 
  \parbox[t]{0.3in}{\includegraphics[width=0.7in]{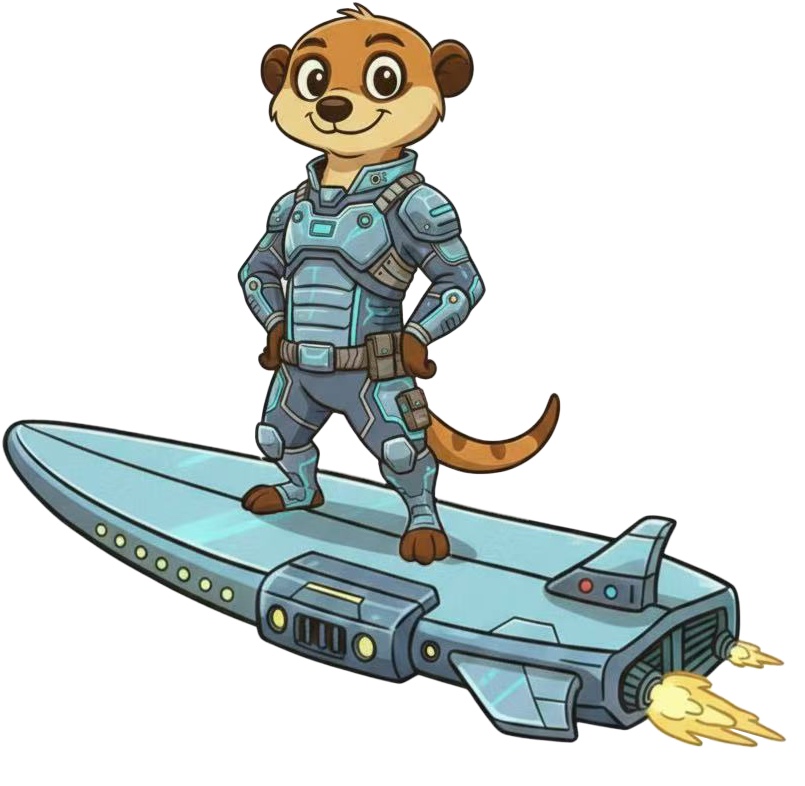}}\quad 
  }
  \begin{tabular}[t]{l}
  \parbox[t]{0.63\textwidth}{ 
    Scaling Agents via Continual Pre-training
  }
  \end{tabular}
}

\author{%
{
Liangcai Su$^{*}$, Zhen Zhang$^{*}$, Guangyu Li$^{*}$, Zhuo Chen$^{*}$, Chenxi Wang$^{*}$, Maojia Song, Xinyu Wang$^{(\textrm{\Letter})}$\thanks{Equal Contributions. Xinyu Wang is the project leader.}, Kuan Li, Jialong Wu, Xuanzhong Chen, Zile Qiao, Zhongwang Zhang, Huifeng Yin, Shihao Cai, Runnan Fang, Zhengwei Tao, Wenbiao Yin,  Chenxiong Qian, Yong Jiang$^{(\textrm{\Letter})}$, Pengjun Xie, Fei Huang, Jingren Zhou 
}
\\
Tongyi Lab\symboletongyi, Alibaba Group
}

\newtcbox{\clue}{on line, colback=yellow!25, colframe=black, boxrule=0.5pt, arc=2pt, boxsep=1pt, left=2pt, right=2pt, top=1pt, bottom=1pt}

\begin{document}

\maketitle


\begingroup
  \renewcommand\thefootnote{\Letter}  
  \footnotetext{ Corresponding author. \{tomas.wxy, yongjiang.yj\}@alibaba-inc.com
  } 
\endgroup


\begin{abstract}
Large language models (LLMs) have evolved into agentic systems capable of autonomous tool use and multi-step reasoning for complex problem-solving. However, post-training approaches building upon general-purpose foundation models consistently underperform in agentic tasks, particularly in open-source implementations. We identify the root cause: the absence of robust agentic foundation models forces models during post-training to simultaneously learn diverse agentic behaviors while aligning them to expert demonstrations, thereby creating fundamental optimization tensions. To this end, we are the first to propose incorporating Agentic Continual Pre-training \textbf{(Agentic CPT)} into the deep research agents training pipeline to build powerful agentic foundational models. Based on this approach, we develop a deep research agent model named AgentFounder. We evaluate our AgentFounder-30B on 10 benchmarks and achieve state-of-the-art performance while retains strong tool-use ability, notably \textbf{39.9\%} on BrowseComp-en, \textbf{43.3\%} on BrowseComp-zh, and \textbf{31.5\%} Pass@1 on HLE.

\end{abstract}
\begin{figure}[h]
    \centering
    \includegraphics[width=0.95\textwidth]{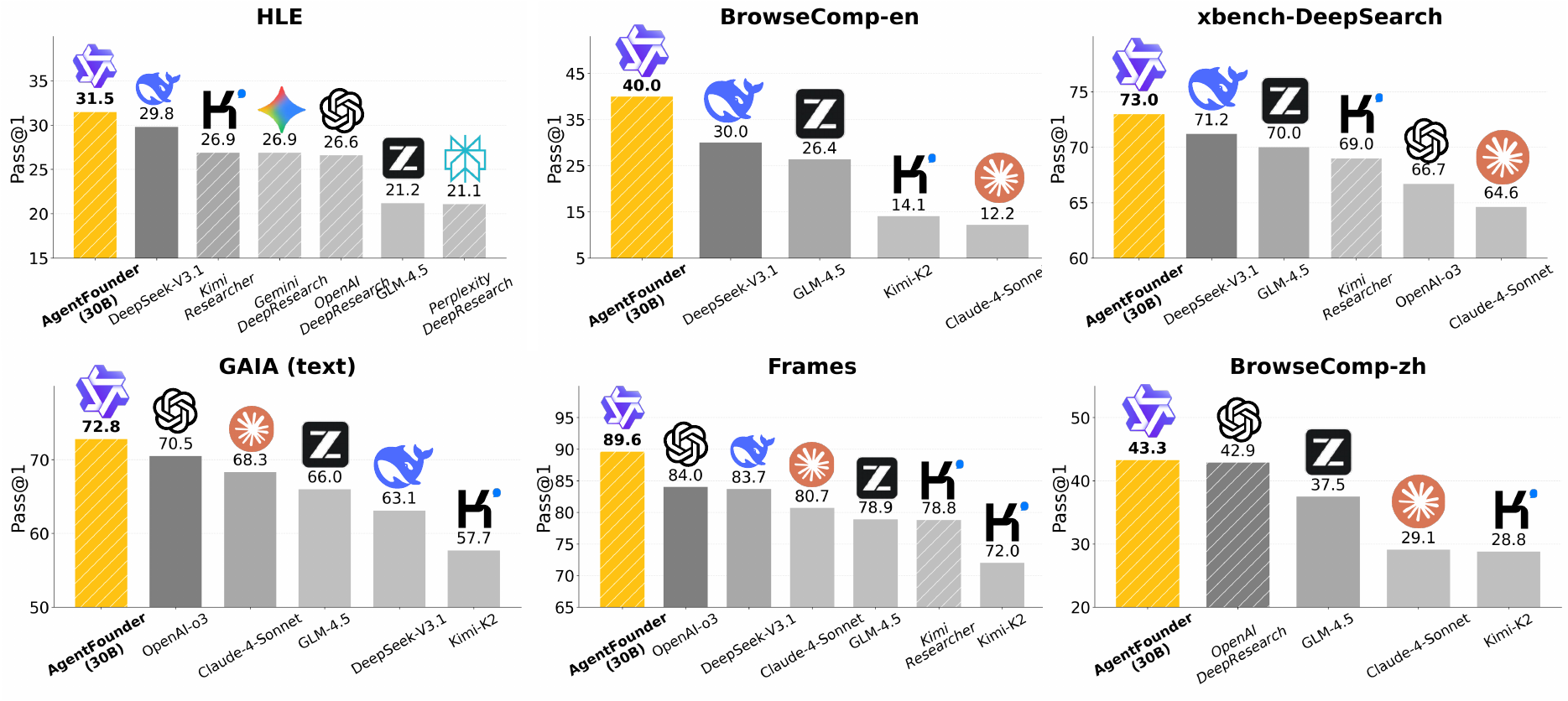}
    \caption{Performance comparison between AgentFounder and state-of-the-art deep research agents.}
    \label{fig:main}
\end{figure}

\section{Introdcution}
The prevailing understanding of \textit{alignment} in language models has been shaped by static interaction paradigms, which focus on ensuring model outputs align with human preferences \citep{RLHF2022}. However, as models evolve into autonomous agents capable of multi-step reasoning, dynamic tool invocation, and complex environmental interactions \citep{yao2023react, toolformer, tool_learning_survey_2024}, this definition requires fundamental expansion. To address this paradigm shift, we introduce \textit{agentic alignment}, which requires language models to maintain behavior consistency with human expert demonstrations when solving complex tasks in dynamic environments. This encompasses reasoning chains, tool invocations, and adaptive responses to unpredictable environmental changes (e.g., tool failure, misleading information). Further, language models achieving such alignment for web retrieval and knowledge-intensive tasks can be considered \textit{deep research agents}, capable of autonomously orchestrating sophisticated workflows through search, browsing, code execution and other tools to accomplish complex tasks and provide trustworthy answers, e.g., OpenAI Deep Research \citep{openai2025deepresearch}.

Unfortunately, when extended to agentic alignment, current post-training methods, including Supervised Fine-Tuning (SFT) and Reinforcement Learning Fine-Tuning (RL) \citep{flanv2, RLHF2022, taori2023stanford}, demonstrate limited effectiveness, particularly in open-source implementations. Even leading agentic models such as WebSailor (12.0) \citep{li2025websailor}, GLM-4.5 (26.4) \citep{zeng2025glm}, and DeepSeek-V3.1 (30.0) \citep{deepseekai2025deepseekv3_1} exhibit substantial performance gaps compared to OpenAI's Deep Research (51.5) on challenging benchmarks like BrowseComp \citep{bc_en}. A potential explanation is that these methods predominantly rely on general-purpose foundation models\footnote{GLM-4.5 incorporates synthetic agent trajectories during mid-training.}, such as Qwen2.5-72B \citep{qwen2.5}, which presents a critical bottleneck. Specifically, deep research agents require traversing vast policy spaces where SFT's reliance on complete, high-quality trajectory data makes comprehensive coverage infeasible. Moreover, agent trajectories are inherently long and complex, making precise definitions of "correct behavior" challenging. Consequently, both SFT and RL training depend on limited deterministic supervisory signals that lock models into replicating specific behavioral patterns rather than develop flexible decision-making capabilities. Fundamentally, general-purpose foundation models lack agentic inductive biases, forcing post-training to simultaneously learn capabilities and alignment, creating inherent optimization conflicts. Crucially, pathways toward developing agentic foundation models themselves remain largely unexplored.



Thus, beyond post-training, we redefine the agentic alignment training pipeline by introducing Agentic Continue Pre-training (\textbf{Agentic CPT}) as an intermediate scaling layer for agentic alignment. The core objective of Agentic CPT is to deliver a pre-aligned agentic foundation model that naturally supports agentic behaviors for effective downstream fine-tuning. To achieve the goal, agentic CPT operates on two fundamental principles. First, during the data collection phase, seed data sources must be broad and not confined to any single domain. Second, when preparing training data, the training data must comprehensively include various types of agentic behaviors, preventing models from imitating and memorizing specific behavioral patterns that would compromise their behavioral exploration capabilities.




Guided by these principles, we develop AgentFounder, a deep research agent model obtained through Agentic CPT and subsequent post-training, starting from Qwen3 series models \citep{yang2025qwen3}. Our agent relies on a systematic and scalable data synthesis approach, comprising First-order Action Synthesis (FAS), Higher-order Action Synthesis (HAS), and a two-stage training strategy.

For first-order action synthesis, we construct (question, planning, action) data tuples by reorganizing diverse knowledge sources into entity-knowledge mappings and randomly sampling entities to create diverse QA pairs. Then, based on QAs, we consider two types of action synthesis: planning data that generates reasonable planning and next-step tool invocations, and logical reasoning data that produces step-by-step reasoning processes leading to final answers. For higher-order action synthesis, we remodel trajectories as a multi-step decision-making process by expanding potential exploration paths at each step, including reasoning and tool invocation. We then merge trajectories with newly generated exploration paths into multi-step decision-making text data. This approach allows models to explore various feasible behavioral patterns, enhancing their exploration and decision-making capabilities. Importantly, both synthesis approaches operate without external tool invocations, enabling large-scale data generation in offline environments without API costs.

To efficiently absorb these two types of synthesis agentic data, we propose a progressive two-stage training strategy. The first stage primarily utilizes FAS data and short HAS data within a 32K context window, while the second stage focuses on high-quality HAS data with an extended 128K context length.

Then, we evaluate AgentFoudner-30B against state-of-the-art (SOTA) models including general LLMs with tools, commercial and open-source deep research agents across 10 benchmarks. AgentFoudner-30B achieves superior performance, obtaining \textcolor[RGB]{103,49,226}{\textbf{39.9\%}} on BrowseComp-en, \textcolor[RGB]{103,49,226}{\textbf{43.3\%}} on BrowseComp-zh, \textcolor[RGB]{103,49,226}{\textbf{72.8\%}} on GAIA, \textcolor[RGB]{103,49,226}{\textbf{31.5\%}} on HLE, and \textcolor[RGB]{103,49,226}{\textbf{73.0\%}} on xbench-DeepSearch, achieving new SOTA results. Remarkably, scaling AgentFounder's training data yields steady performance average gains across all benchmarks, exhibiting promising scaling law behaviors. Besides, we find that AgentFounder still maintains strong general tool-use capabilities, suggesting potential for general-purpose agents in the future.

\section{AgentFounder: Agentic Foundation Model}
This section introduces AgentFounder, our proposed deep research agent that leverages agentic continual pre-training to create pre-aligned agentic foundation models for downstream fine-tuning. We focus on detailing how AgentFounder implements Agentic CPT, specifically covering our systematic and scalable training data synthesis methods and progressive training strategies  that prepare foundation models for effective downstream fine-tuning.
\subsection{Overview of Agentic Training Pipeline}
\begin{figure}[t]
    \centering
    \includegraphics[width=0.9\textwidth]{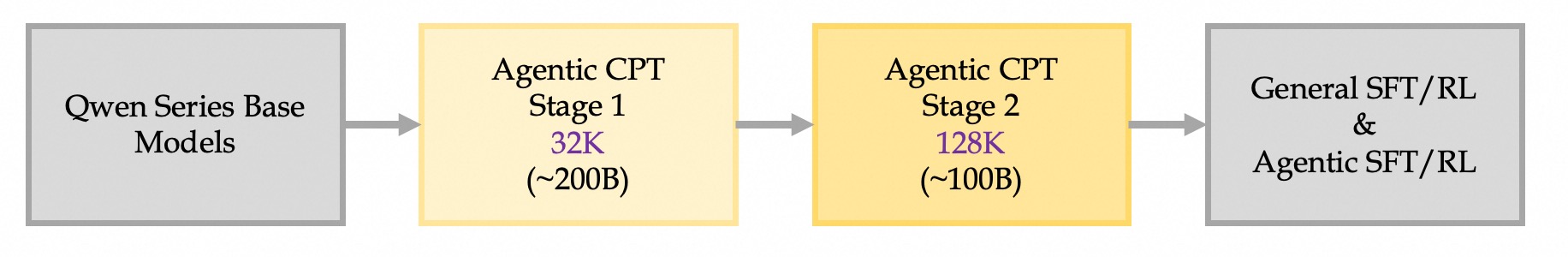}
    \caption{Agentic Training Pipeline.}
    \label{fig:pipeline}
\end{figure}
Contemporary large language model development follows a well-established two-stage paradigm consisting of pre-training and post-training. During pre-training, models acquire broad knowledge from diverse data sources including web-crawled text, code repositories, academic literature, books, and news articles, following the fundamental \textbf{next-token prediction} paradigm with cross-entropy loss:
\begin{equation}
\mathcal{L} = -\sum_{t=1}^{T} \log P(x_{t+1}|x_1, x_2, ..., x_t)~\label{eq:pretraining}
\end{equation}
where $P(x_{t+1}|x_1, x_2, ..., x_t) = \text{softmax}(W_o h_t)$, $h_t$ represents the hidden state at position $t$, and $W_o$ is the output projection matrix.

Post-training typically encompasses SFT and RL, enabling LLMs to follow instructions and align with human preferences. For agentic capabilities, post-training teaches LLMs to utilize tools effectively and perform multi-step reasoning for complex tasks. To build AgentFounder, we conduct supervised fine-tuning using a strategically proportioned mixture of general instruction data and agent trajectory demonstrations to ensure the LLM develops both general-purpose capabilities and specialized agentic abilities for deep research agents.

Beyond the standard pipeline, we fundamentally redesign the training paradigm by integrating Agentic Continual Pre-Training as an additional stage between pre-training and post-training, as illustrated in Figure~\ref{fig:pipeline}. Starting from Qwen's pre-trained foundation models (\textit{e.g.}, Qwen3-30B-A3B-Base\footnote{\url{https://huggingface.co/Qwen/Qwen3-30B-A3B-Base}}), our enhanced training pipeline consists of:

\begin{itemize}
    \item Agentic CPT Stage 1: We process approximately 200B tokens of agent data and knowledge reasoning corpora with 32K context length, following the same next-token prediction paradigm as Eq.~\ref{eq:pretraining}. This stage enables the preliminary acquisition of agentic behaviors including tool invocation patterns and multi-step reasoning chains.
    \item Agentic CPT Stage 2: We further refine these capabilities using ~100B tokens of carefully curated, high-quality agent data with extended 128K context windows, allowing the LLM to develop a sophisticated understanding of complex action spaces and long-horizon planning strategies.

\end{itemize}

\begin{figure}[t]
    \centering
    \includegraphics[width=\textwidth]{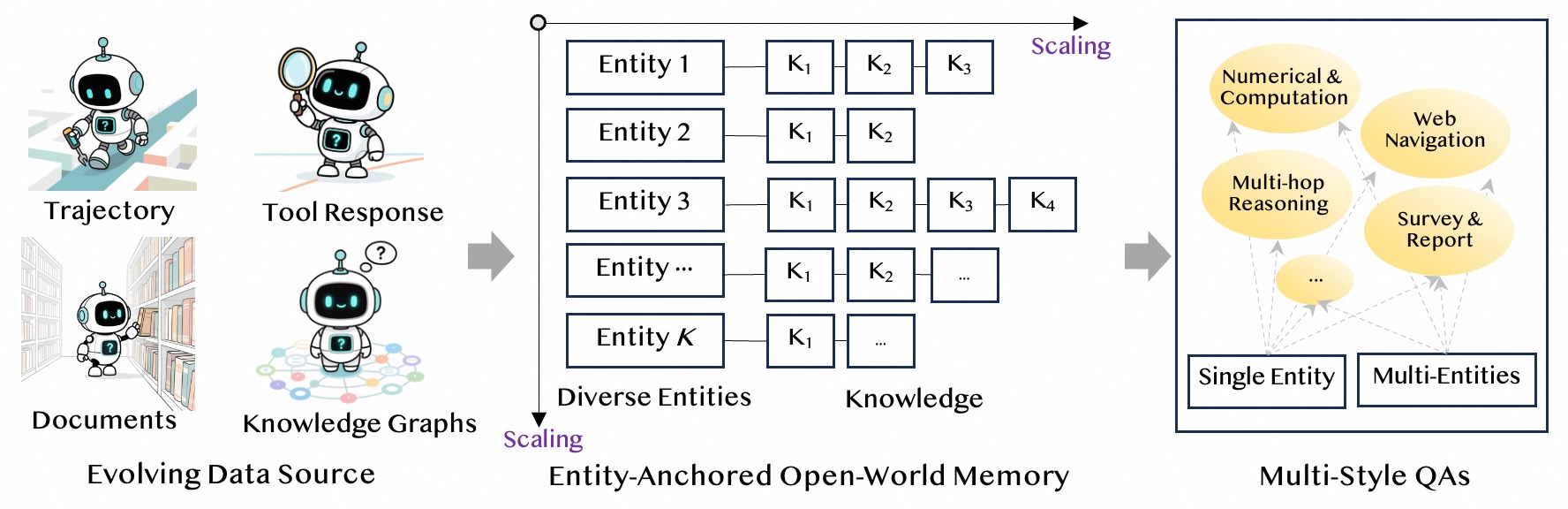}
    \caption{Multi-Style Question-Answer Generation Based on Scalable Information Sources.}
    \label{fig:multi_qa}
\end{figure}

\subsection{First-order Action Synthesis with Zero Supervisory Signal}
In this section, we present an agentic data synthesis methodology, First-order Action Synthesis (\textbf{FAS}), which operates without supervisory signals and relies solely on diverse data sources. FAS consists of contextual scenario construction and two types of action synthesis: planning actions and reasoning actions.

\subsubsection{Scaling Training Contexts via Knowledge-to-Question Transformation}\label{sec:training_contexts}
Conventional continual pre-training focuses on knowledge adaptation, particularly domain-specific knowledge acquisition \citep{shi2024continual}. In contrast, Agentic CPT targets the adaptation of agentic capabilities, which are domain-agnostic abilities that transcend specific domains and enable universal tool utilization and multi-step reasoning. Since these abilities must function effectively across diverse application scenarios, this capability adaptation necessitates training data spanning multiple domains to ensure broad transferability and applicability of the acquired skills.

To achieve this scale and diversity, we collect multiple data types: discarded trajectories from post-training datasets, historical tool invocation results (\textit{e.g.}, search queries and responses), and publicly available corpora (\textit{e.g.}, CommonCrawl \citep{commoncrawl2025}). These sources provide continuous scalability while maintaining comprehensive domain coverage. For instance, search results can be persistently extracted from trajectories generated during multiple rollouts in the reinforcement learning process.


However, while trajectory data directly demonstrates agentic behavior, the remaining sources primarily contain static knowledge. To maximize the utility of this static knowledge for agentic capability training, we reformulate it into diverse training contexts that simulate real-world agent scenarios. This is accomplished through a two-phase transformation approach that systematically generates multi-style questions from static knowledge sources (illustrated in Figure~\ref{fig:multi_qa}):

\textbf{Phase 1: Entity-Anchored Open-World Knowledge Memory.} 
We transform continuously updated unstructured text from various sources into an open-world memory, where entities serve as indexing keys mapping to their associated declarative statements. Unlike traditional knowledge graphs with fixed schemas~\citep{auer2007dbpedia,vrandevcic2014wikidata}, we do not focus on inter-entity relationships, but instead enhance the density of corresponding knowledge statements through reformulation, preserving critical information such as temporal markers, sources, and original stylistic features. For instance, web data containing ``The number of tourist arrivals in France increased from 3,793 thousand in May 2025 to 4,222 thousand in June'' can be reformulated as: (``France'', ``Tourist arrivals in France reached 4,222 thousand in June 2025''), rather than limiting to conventional wiki-style knowledge such as ``Paris is the capital of France.'' Through continuous updates from search results and web access outcomes, both entities and their corresponding knowledge statements continuously expand, forming a living memory system whose content better aligns with the information distribution of the internet world.

\textbf{Phase 2: Multi-Style Question Synthesis.} 
Drawing on the entity-anchored open-world memory, we sample entity clusters along with their associated knowledge statements to synthesize diverse questions spanning factual retrieval, numerical computation, multi-hop reasoning, and synthesis tasks. This transformation converts static knowledge into dynamic problem-solving contexts that necessitate active information retrieval, integration, and tool innovations, thereby establishing a foundation for reinforcing agentic behaviors in subsequent data synthesis. By exploiting the high density of statements per entity to induce implicit cross-entity links, our approach yields diverse, reliable, and novel questions. Unlike WebSailor~\citep{li2025websailor} which requires explicit relationship construction between entities, our method leverages the richness of reformulated knowledge statements from the entity-anchored memory to create natural knowledge intersections, substantially improving the reliability and novelty of question generation. Moreover, the comprehensive knowledge coverage per entity enables synthesis of sophisticated questions even from single-entity contexts.

\begin{tcolorbox}[
  colback=mylightpurple,   
  colframe=mydarkpurple,   
  coltitle=white,          
  fonttitle=\bfseries,     
  title=Example: Generated Questions with the entity "Paris",
  enhanced,
  boxrule=0.8pt,           
  arc=2mm,                 
  left=2mm, right=2mm, top=1mm, bottom=1mm
]
\textbf{Source Entity:} Paris
\noindent

\textbf{Knowledge 1:} \emph{The Louvre welcomed 8.7 million visitors in 2024. Visitor numbers thus remained at 2023 levels (8.9 million visitors) in the rather unique context of the Paris 2024 Olympic and Paralympic Games
}

\noindent
\textbf{Knowledge 2:} \emph{In 2023, France’s bedbug outbreak sparked a political row; Paris pushed pre-2024 Olympic action, and the transport minister summoned transit operators.}

\noindent
\textbf{Knowledge 3:} \emph{At Paris 2025, Airbus announced 132 firm orders and up to 106 additional options/increases—AviLease (40+37), Riyadh Air (25+25), ANA (27), LOT (40+44)—underscoring Paris as a global aviation deal hub.
}

\dots


\textbf{Question:} At the biennial aerospace marketplace named after the city whose pyramid-fronted museum recorded high single-digit millions of visitors during a period of global athletic celebration, and where the year before a citywide nuisance led authorities to convene transit operators, which buyer placed a perfectly balanced commitment with firm orders equal to options?

\noindent

\textbf{Answer:} Riyadh Air

\end{tcolorbox}

This example synthesizes a question from the entity Paris using three news-sourced statements. Knowledge 1 offers an indirect locational anchor via visitor counts at the pyramid-fronted museum during a global sporting period, which localizes the marketplace without naming the city. Knowledge 2 adds a prior-year civic disturbance that supplies a relative timeline and helps isolate the 2025 iteration of the biennial event. Knowledge 3 provides the order breakdown at that edition, establishes the criterion “firm orders equal options,” and narrows the answer to Riyadh Air. Since these facts are recent and fluid, reliable resolution typically requires external retrieval with search tools.

\subsubsection{Planning Action Synthesis}
\begin{figure}[tb]
    \centering
    \includegraphics[width=0.6\textwidth]{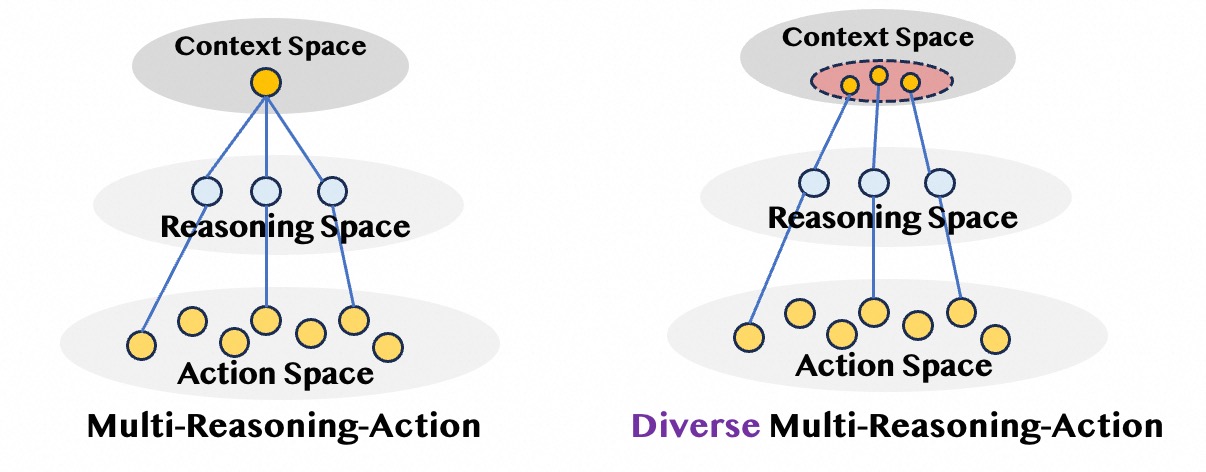}
    \caption{Planning Action Synthesis.}
    \label{fig:mral}
\end{figure}

While we have constructed large-scale, scalable questions to cover diverse scenarios, questions alone cannot constitute complete agent training data, because corresponding reasoning processes and action sequences are also required.

\textbf{Scalability Challenges.} The intuitive solution is to leverage LLMs to perform tool invocations and reasoning for each question, generating complete agent trajectories. However, this approach faces severe scalability bottlenecks: (1) commercial API costs are prohibitively expensive, particularly for search engine APIs (\textit{e.g}., Google Search API) and web access APIs (\textit{e.g}., Jina Reader API \citep{jina} and Scraper \citep{scraperapi_website}); (2) complete trajectory generation is inefficient and cannot meet the massive data requirements for continual pre-training.


\textbf{Insighs.} To address this challenge, we propose a scalable reasoning-action synthesis approach. We observe that the initial analysis of complex problems by LLMs typically involves problem decomposition, information requirement identification, and solution planning, which inherently constitutes high-quality planning data. More importantly, \textbf{the quality of first-step reasoning exhibits strong positive correlation with final task completion rates.}  

Based on this finding and inspired by multi-reference learning~\citep{zheng2018multi,banerjee2005meteor}, FAS generates multiple reasoning-action data for each problem in planning action synthesis. As illustrated in Figure~\ref{fig:mral} (left), given a query $Q$, we employ LLMs to generate $K$ diverse problem analyses along with their corresponding first-step action predictions (tool invocations or direct answers). This approach yields two key advantages: (1) it generates only reasoning chains and tool calls without incurring actual API invocation costs during training data production; (2) the $K$ distinct analytical perspectives effectively expand the action space exploration for each problem.



However, we realize that the aforementioned approach faces two limitations: (1) the generated reasoning-action data may still exhibit similarity despite adjusting parameters like temperature to enhance diversity; (2) it leads to repetitive question text in training data, which is not our optimization target. To address these issues, we propose an improved strategy. Instead of generating $K$ iterations of reasoning-action data for a single question, we generate reasoning-action data for $K$ different questions that share the same knowledge memory but differ in style. This approach better covers training contexts and explores the potential reasoning-action space more comprehensively. In practice, we adopt this question-level diversity expansion as the planning action synthesis method in FAS.

\textbf{Reject Sampling with Knowledge Alignment Verification.} The reasoning and action predictions are not always accurate, and we lack end-to-end trajectory signals to verify their correctness. Beneficially, since the knowledge involved in question construction is accessible, we can implement reject sampling based on knowledge alignment verification. Specifically, we employ an LLM-as-Judge approach to evaluate whether the current reasoning and actions have a high probability of acquiring the required knowledge. We find that this rejection sampling strategy effectively filters out the majority of low-quality data, as detailed in Section~\ref{sec:fas_rej}.


    

\subsubsection{Reasoning Action Synthesis}

In the workflow of deep research agents, a critical scenario emerges when comprehensive information has been acquired through various tool invocations or user interactions: how to effectively synthesize this information to generate final answers or reports. Unlike conventional mathematical numerical reasoning, this type of reasoning typically involves logic-based inference grounded in factual information, requiring a balance between divergent thinking and convergent thinking guided by contextual clues, while being difficult to verify through formal methods. To enhance agent capabilities in this domain, we generate synthetic data based on the Question-Answer pairs constructed in Section~\ref{sec:training_contexts}.

Questions constructed from knowledge clusters not only provide the questions itself but also enable precise mapping to the information required for solving them. Based on this foundation, we propose a two-step logical reasoning data synthesis scheme:
\begin{enumerate}
    \item \textbf{Step 1:} We require the large language model to decompose the question $Q$ into multiple sub-questions, then leverage its internal knowledge to generate reasonable speculations and answers for each sub-question, producing preliminary answer $A_1$.
    \item \textbf{Step 2:} Given the question $Q$ and its mapped requisite knowledge, we require the model to refine answer $A_1$, correct logical errors, and generate the final answer $A_2$.
\end{enumerate}

Notably, we prohibit the model from invoking any external tools during both stages. The motivation behind this two-step design is that if we directly provide both the question and necessary knowledge, the model tends to mechanically utilize the given knowledge as intermediate reasoning nodes rather than simulating an authentic thinking process. We show an example below.

\begin{tcolorbox}[
  breakable,
  colback=mylightpurple,   
  colframe=mydarkpurple,   
  coltitle=white,          
  fonttitle=\bfseries,     
  title=Example: An Overview Reasoning Action Synthesis Data,
  enhanced,
  boxrule=0.8pt,           
  arc=2mm,                 
  left=2mm, right=2mm, top=1mm, bottom=1mm
]
\textbf{\textcolor[RGB]{65,105,225}{(Clue1: The Location)}}: First, the question mentions ..., which presents several potential candidates: the Louvre in Paris with its iconic glass pyramid entrance, .... 
However, the crucial filtering criterion is the high visitor numbers "during a period of global athletic celebration." This could refer to various major international sporting events: .... 
Among these candidate cities, only Paris simultaneously possesses both a pyramid-fronted museum and hosted a global athletic celebration, ...., \textbf{thus confirming this refers to the Louvre during the 2024 Paris Olympics.} 

\textbf{\textcolor[RGB]{65,105,225}{(Clue2: The Time)}}: Next, the question describes "the year before a citywide nuisance led authorities to convene transit operators," requiring consideration of events in 2023 that could have affected all of Paris and involved the transportation system. Possible candidates include massive public transit strikes, severe air pollution episodes, .... 
Having established the city as Paris, the "biennial aerospace marketplace named after the city" clearly points to the Paris Air Show, which is held in odd-numbered years, \textbf{thus locking in the timeframe as the 2025 Paris Air Show.}

\textbf{\textcolor[RGB]{65,105,225}{(Clue3: The Order)}}: Finally, the question asks which buyer placed "a perfectly balanced commitment with firm orders equal to options." Examining the major order announcements at the 2025 Paris Air Show reveals: .... Therefore, the answer is \textbf{Riyadh Air}.

\end{tcolorbox}

\textbf{Reject Sampling.} Furthermore, we employ an LLM-as-judge approach to evaluate the alignment between the generated answer $A_2$ and the ground truth answer, implementing rejection sampling accordingly. If the final answer proves correct, we consider the reasoning process contained in $A_2$ to be reliable. Through this methodology, we generate a substantial corpus of high-quality logical reasoning chain-of-thought data. It is worth emphasizing that this logical deduction capability constitutes a fundamental competency required by deep research agents throughout the entire problem-solving lifecycle.

\subsection{High-order Action Synthesis with Supervisory Signal}
\textbf{Trajectory-Reuse Challenging.} During the post-training phase of agent models, both reject sampling fine-tuning and RL generate substantial volumes of trajectory data. However, these methods rely heavily on trajectory-level delayed feedback for quality assessment, which results in numerous trajectories being either discarded entirely or utilized only once when they fail to meet stringent quality thresholds. This coarse-grained evaluation approach leads to significant waste of the learning signals embedded within real trajectories. While step-level evaluation could theoretically provide better leverage of these signals, precisely assessing intermediate steps remains challenging. Naively incorporating such uncertain reward signals into SFT or RL training risks model collapse. Therefore, effectively reusing sub-optimal trajectories while maintaining training stability remains a key challenge.

\textbf{Insights.} We recognize that each step in a trajectory is supported by high-quality context including the original question, prior steps, and their real feedback. This context defines a distinct reasoning state with a broad space of feasible reasoning–action options. Thus, every step is fundamentally a hidden decision process. However, although agents often generate multiple candidates within a single reasoning–action turn (\textit{e.g.}, alternative queries or exploration directions), these candidates remain internal branches of the same path, and supervision mainly rewards reproducing the full trajectory. Consequently, models learn to imitate a sequence rather than to perform decision-making at critical steps. We therefore shift the objective from trajectory imitation to step-wise decision-making, explicitly exploiting the choice space at each step to move from trajectory reproduction to decision-based reasoning.

\begin{figure}[t]
    \centering
    \includegraphics[width=0.99\textwidth]{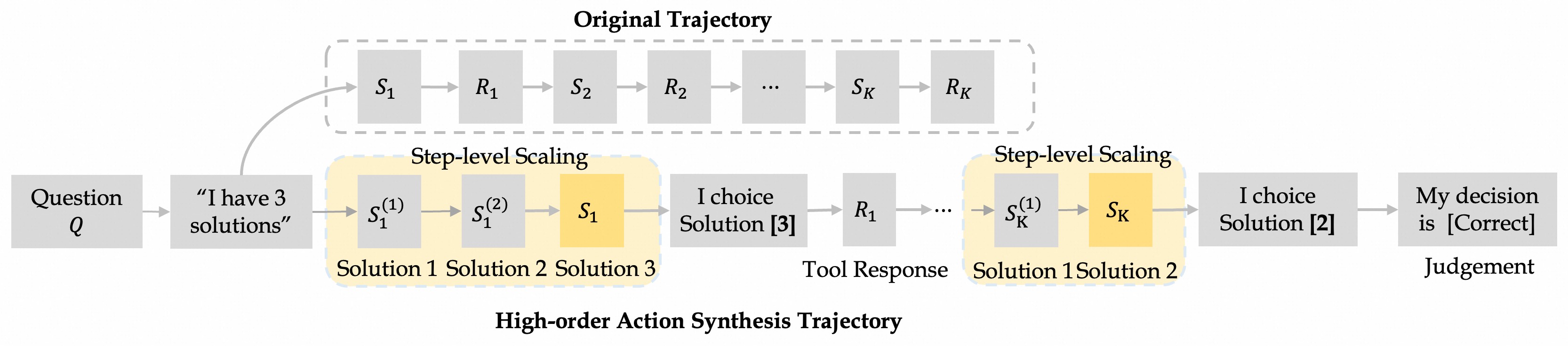}
    \caption{Comparison of high-orider action synthesis data and the original trajectory. }
    \label{fig:mr}
\end{figure}

To this end, we introduce \textbf{High-order Action Synthesis (HAS)}, a multi-decision action synthesis method through step-level scaling. The core idea is to expand the reasoning-and-action option set at each step to fully explore the local action space, then transform the original trajectory and explored reasoning-action space into decision processing with feedback. Specifically, given a problem $Q$ and an agent trajectory $T = \{(S_1, R_1), \ldots, (S_K, R_K)\}$, where $S_k$ represents the ``reasoning and tool invocation'' at the $k$-th step and $R_k$ denotes the corresponding tool/environment response, the entire trajectory carries a binary judgment $J \in \{0, 1\}$ indicating failure and success respectively. HAS comprises two components:


\textbf{(1) Step-level Scaling.} For any step $S_k$, we denote its conditional context as $C_k = (Q, S_1, R_1, \ldots, S_{k-1}, R_{k-1})$. Without actual tool execution, we employ LLMs to generate $N$ alternative ``thought and invocation'' candidates for context $C_k$: $A_k = \{S_k^{(1)}, \ldots, S_k^{(N)}\}$. We merge the {original step} $S_k^{(0)} \equiv S_k$ with these candidates to obtain $N+1$ feasible steps, then {randomly shuffle} them to form sequence $\tilde{A}_k$ while recording the original step's position $n_k$ in the sequence. This expansion explores different decision possibilities at the cognitive level, enriching the original trajectory into a decision space with  $(N+1) \times K$ potential reasoning-actions.



\textbf{(2) Contrastive Decision-Action Synthesis.} We transform the trajectory with expanded options into a progressive decision-making process. As illustrated in Figure~\ref{fig:mr}, starting from problem $Q$ for each step, we explicitly simulate a multi-option selection and decision process. For the $k$-th step, we enumerate each option in $\tilde{A}_k$ and insert a local action decision statement: ``I will choose option $n_k$'', immediately followed by the corresponding real response $R_k$. Finally, we append the judgment text: ``My decision is \{Correct/Incorrect\}'' (corresponding to $J$). The complete synthetic training sample is obtained by concatenating the problem, the choice-decision process for each step, and the final judgment text.

This approach circumvents the risks associated with directly using uncertain step-level rewards while enabling the model to learn from diverse reasoning paths, thereby preventing overfitting to specific trajectory patterns. Through this synthesis strategy, previously underutilized trajectory data is transformed into rich training signals, significantly improving the sample efficiency of the agentic learning process.


\section{Experiments}
In this section, we conduct comprehensive empirical studies to validate the performance of AgentFounder. Our experimental design addresses the following research questions:
\begin{itemize}
    \item \textbf{RQ1:} How does AgentFounder compare against existing state-of-the-art deep research agents?
    \item \textbf{RQ2:} Can agentic CPT effectively enhance diverse post-training methods for agentic alignment?
    \item \textbf{RQ3:} What is the effectiveness of our proposed two-stage training strategy in the agentic CPT?
    \item \textbf{RQ4:} Which data type (i.e., FAS, HAS) contributes more effectively to agentic CPT?
    \item \textbf{RQ5:} Do scaling laws apply to both data volume and model size in agentic CPT?
\end{itemize}

\subsection{Experimental Setup}
\subsubsection{Data} \label{sec:exp_data}
\textbf{Agentic CPT Data.} Our continual pre-training corpus comprises a carefully curated heterogeneous mixture of: (1) high-quality web-crawled data filtered for factual accuracy, (2) historical tool invocation records, e.g., search results and web page content (3) offline Wikipedia data, and (4) mixed-quality discarded trajectories from previous post-training iterations.

\textbf{Post-training Data.}  
To better unlock the potential of our foundation models and demonstrate the adaptability of AgentFounder-Base across different post-training strategies, we employ three distinct SFT configurations:
\begin{itemize}
    \item \textbf{SFT-A:} Employs a two-stage training paradigm, first on general conversational data, followed by specialized React-style agent trajectories with explicit reasoning chains.
    \item \textbf{SFT-B:} An enhanced version of SFT-A that maintains the two-stage training paradigm but incorporates a balanced mixture of general conversational data and React-style trajectories in each stage.
    \item \textbf{SFT-C:} Employs a two-stage training paradigm with general conversational SFT data and React with summarized reasoning trajectories.
\end{itemize}
For all SFT configurations, the set of the challenging information-seeking questions is constructed following the methodology from WebSailor-V2 \citep{li2025websailor-v2}, WebResearcher \citep{qiao2025webresearcher}, WebWeaver \citep{webweaver} and AgentScaler \citep{agentscaler}.


\subsubsection{Baselines}
We evaluate against three categories of strong agent models:
\begin{itemize}
    \item \textbf{General LLMs with tools}: Qwen3-30B-A3B-2507 \citep{yang2025qwen3}, Qwen3-235B-A22B-2507 \citep{yang2025qwen3}, DeepSeek-R1-0528 \citep{r1} and Claude-4-Sonnet \citep{anthropic2025claude4}.
    \item \textbf{Commercial deep research agents}: Kimi-Researcher \citep{team2025kimi}, OpenAI-o3 \citep{o3}, OpenAI Deep Research \citep{openai2025deepresearch}, Grok Deeper Search \citep{grok}, Perplexity Deep Research \citep{perplexity2025deep}, Gemini Deep Research \citep{google2025gemini_deep_research}.
    \item \textbf{Open-source deep research agents}: WebThinker-32B-RL \citep{li2025webthinker}, ASearcher-Web-QwQ \citep{gao2025beyond}, WebSailor-72B \citep{li2025websailor}, WebShaper-72B \citep{tao2025webshaper}, AFM-32B-RL \citep{afm}, MiroThinker-32B-DPO$_\text{v0.2}$ \citep{2025mirothinker}, DeepDiver-V2-38B \citep{deepdiver2}, WebExplorer-8B \citep{liu2025webexplorerexploreevolvetraining}, DeepDive-32B \citep{deepdive}, Kimi-K2-Instruct \citep{team2025kimi}, GLM-4.5 \citep{zeng2025glm}, and DeepSeek-V3.1 \citep{deepseekai2025deepseekv3_1}.
\end{itemize}
We prioritize official results reported by model providers or benchmark organizers, or scores reported by other published works, and evaluate remaining models under our standardized setup. If a work provides multiple agent models, we report only the strongest one.



\subsubsection{Benchmarks}

We evaluate across two categories of benchmarks to comprehensively assess agent capabilities:  

\textbf{General web search benchmarks:} \textit{BrowseComp-en} \citep{bc_en}, \textit{BrowseComp-zh} \citep{bc_zh}, \textit{GAIA}\footnote{We use the text-only subset consisting of 103 questions.} \citep{mialon2023gaia}, \textit{Xbench-DeepSearch} \citep{xbench} and \textit{WebWalkerQA} \citep{webwalker}, targeting general-purpose browsing, search and reasoning tasks. 

\textbf{Scenario-targeted web search benchmarks:} This category encompasses specialized evaluations designed for realistic task settings across diverse domains. 
\begin{itemize}
    \item \textit{DeepResearch Bench} \citep{du2025deepresearch} assesses comprehensive research report generation through expert-level tasks across multiple academic fields.
    \item  \textit{SEAL-0} \citep{pham2025sealqa} evaluates model robustness when facing conflicting or misleading search results. 
    \item \textit{Frames} \citep{frames} evaluates a model’s ability to conduct multi-perspective reasoning and role-based information synthesis, requiring consistent integration of evidence across different contextual frames.
    \item \textit{HLE} (Humanity's Last Exam) \citep{phan2025humanity} evaluates models on expert-level questions across diverse subjects.
    \item \textit{Academic Browse} \citep{zhou2025academicbrowse} focuses on scholarly research capabilities such as literature navigation and knowledge synthesis.
\end{itemize}



\subsubsection{Evaluation Protocol}
\textbf{Tools}. By default, evaluated models are equipped with five core tools: \texttt{Search} (web search with result ranking), \texttt{Visit} (webpage content extraction), \texttt{Google Scholar} (academic literature access), \texttt{Python Interpreter} (code execution), and \texttt{File Parser} (document processing). More details are presented in Appendix \ref{apx:tool}.


\textbf{Hyper-Parameters}. When evaluating our AgentFounder models, we employ specific inference parameters to ensure stable and reproducible results: temperature 0.85, repetition penalty 1.1, and top-p 0.95. These settings are recommended based on extensive empirical validation to optimize the balance between creativity and consistency in agentic reasoning tasks. We set a maximum tool usage limit of 128 calls per task and constrain the context length to 128K tokens.

\subsection{Performance Comparison between Agentic Models (RQ1)}

\begin{table}[t]
    \caption{Results on general web search benchmarks. $^\dagger$ indicates results reported in official sources or prior work.}
    \centering
    \resizebox{\textwidth}{!}{
    \begin{tabular}{l|c|c|c|c|c}
    \toprule
    \textbf{Backbone} & \textbf{BrowseComp-en} & \textbf{BrowseComp-zh} & \textbf{GAIA} & \textbf{Xbench-DeepSearch} & \textbf{WebWalkerQA} \\
    \midrule
    \rowcolor[RGB]{229,229,252}\multicolumn{6}{c}{\emph{\textbf{General LLMs with tools}}} \\
    \midrule
    Qwen3-30B-A3B & 0.5 & 13.5 & 35.9 & 32.0 & 46.9 \\
    Qwen3-235B-A22B & 2.3 & 29.4  & 45.6 & 46.0 & 59.6 \\
    DeepSeek-R1 & 8.9$^\dagger$ & 35.7$^\dagger$&  - & 55.0$^\dagger$ & - \\
    Claude-4-Sonnet & 12.2$^\dagger$ & 29.1$^\dagger$& 68.3$^\dagger$ & 64.6$^\dagger$ & 61.7$^\dagger$ \\
    \midrule
    \rowcolor[RGB]{229,229,252}\multicolumn{6}{c}{\emph{\textbf{Commercial Deep Research Agents}}} \\
    \midrule
    Kimi-Researcher        & -    & -  & -    & 69.0$^\dagger$   & - \\
    OpenAI-o3        & 49.7$^\dagger$     & 58.1$^\dagger$   & 70.5$^\dagger$    & 66.0$^\dagger$   & 71.7$^\dagger$  \\
    OpenAI Deep Research        & 51.5$^\dagger$     & -  & 67.0$^\dagger$    & -  & - \\
    \midrule
    \rowcolor[RGB]{229,229,252}\multicolumn{6}{c}{\emph{\textbf{Open-source Deep Research Agents}}} \\
    \midrule
    WebThinker-32B-RL   & 2.8$^\dagger$   & 7.3$^\dagger$   & 48.5$^\dagger$   & 24.0$^\dagger$   & 46.5$^\dagger$   \\
    ASearcher-Web-QwQ & 5.2$^\dagger$   & 15.6$^\dagger$   & 52.8$^\dagger$   & 42.1$^\dagger$   & 34.3$^\dagger$   \\
    WebSailor-72B   & 12.0$^\dagger$   & 30.1$^\dagger$   & 55.4$^\dagger$   & 55.0$^\dagger$   & - \\
    WebShaper-72B   & -  & -  & 60.1$^\dagger$  & -  & 52.2$^\dagger$ \\
    AFM-32B-RL  & 11.1$^\dagger$  & - & 55.3$^\dagger$  & 63.0$^\dagger$ & -  \\
    MiroThinker-32B-DPO$_\text{v0.2}$   & 17.2$^\dagger$  & 29.4$^\dagger$ & 64.1$^\dagger$  & 56.0$^\dagger$ & 53.6$^\dagger$  \\
    DeepDiver-V2-38B & 13.4$^\dagger$  & 34.6$^\dagger$  & -  & 53.0$^\dagger$  & - \\
    WebExplorer-8B  & 15.7$^\dagger$  & 32.0$^\dagger$  & 50.0$^\dagger$  & 53.7$^\dagger$  & 62.7$^\dagger$ \\
    DeepDive-32B  & 14.8$^\dagger$  & 25.6$^\dagger$  & -  & 50.5$^\dagger$  & - \\
    Kimi-K2 & 14.1$^\dagger$  & 28.8$^\dagger$  & 57.3$^\dagger$  & 50.0$^\dagger$  & 63.0$^\dagger$ \\
    GLM-4.5         & 26.4$^\dagger$  & 37.5$^\dagger$  & 66.0$^\dagger$  & 70.0$^\dagger$  & 65.6$^\dagger$ \\
    DeepSeek-V3.1   & 30.0$^\dagger$  & 49.2$^\dagger$  & 63.1$^\dagger$  & 71.0$^\dagger$  & 61.2$^\dagger$ \\
    \midrule
    \rowcolor[RGB]{229,229,252}\multicolumn{6}{c}{\emph{\textbf{Ours}}} \\
    \midrule
    \textbf{AgentFounder-30B}   & \textbf{39.9}   & \textbf{43.3}  & \textbf{72.8}  & \textbf{73.0}  & \textbf{71.9} \\
    \bottomrule
    \end{tabular}
    }
    \label{tab:general-web-search}
\end{table}

\begin{table}[h]
    \caption{Results on Scenario-Targeted Web Search Benchmarks. $^\dagger$ indicates results reported in official sources or prior work.}
    \centering
    \resizebox{\textwidth}{!}{%
    \begin{tabular}{l|c|c|c|c|c}
    \toprule
    \multirow{2}{*}{\textbf{Backbone}} 
        & \multicolumn{1}{c|}{\textbf{HLE}} 
        & \multicolumn{1}{c|}{\textbf{DeepResearch Bench}} 
        & \multicolumn{1}{c|}{\textbf{Frames}} 
        & \multicolumn{1}{c|}{\textbf{SEAL-0}} 
        & \multicolumn{1}{c}{\textbf{AcademicBrowse}} \\
    & \textbf{Pass@1}
    & \textbf{RACE Overall}
    & \textbf{Pass@1}
    & \textbf{Pass@1}
    & \textbf{Pass@1} \\
    \midrule

    \rowcolor[RGB]{229,229,252}\multicolumn{6}{c}{\emph{\textbf{General LLMs with tools}}} \\
    \midrule
    Qwen3-30B-A3B   & 13.2 & 40.2 & 56.4 & 9.9 & 41.3 \\
    Qwen3-235B-A22B & 20.0 & 44.8 & - & 14.4 & 50.7 \\
    DeepSeek-R1     & 24.8$^\dagger$ & - & 82.0$^\dagger$ & 29.7$^\dagger$ & - \\
    Claude-4-Sonnet & 20.3$^\dagger$ & - & 80.7$^\dagger$  & - & - \\
    \midrule

    \rowcolor[RGB]{229,229,252}\multicolumn{6}{c}{\emph{\textbf{Commercial Deep Research Agents}}} \\
    \midrule
    Grok Deeper Search       & - & 38.2$^\dagger$ & - & - & - \\
    Perplexity Deep Research & 21.1$^\dagger$ & 40.5$^\dagger$ & - & - & - \\
    Gemini Deep Research     & 26.9$^\dagger$ & 49.7$^\dagger$ & - & - & - \\
    Kimi-Researcher          & 26.9$^\dagger$ & 44.6$^\dagger$ & 78.8$^\dagger$ & 36.0$^\dagger$ & - \\
    OpenAI-o3                & 20.2$^\dagger$ & - & 84.0$^\dagger$ & - & - \\
    OpenAI Deep Research     & 26.6$^\dagger$ & 46.5$^\dagger$ & - & - & - \\
    \midrule

    \rowcolor[RGB]{229,229,252}\multicolumn{6}{c}{\emph{\textbf{Open-source Deep Research Agents}}} \\
    \midrule
    ASearcher-Web-QwQ  & 12.5$^\dagger$ & - & 70.9$^\dagger$ & - & - \\
    DeepDive-32B & - & - & 76.1$^\dagger$ & 29.3$^\dagger$ & - \\
    MiroThinker-32B-DPO$_\text{v0.2}$ & 17.8$^\dagger$ & - & 74.8$^\dagger$ & - & - \\
    WebExplorer-8B & 17.3$^\dagger$ & - & 75.7	$^\dagger$ & - & - \\
    Kimi-K2        & 18.1$^\dagger$ & 25.4 & 72.0$^\dagger$ & 25.2 & 47.3 \\
    GLM-4.5        & 21.2$^\dagger$ & 39.2 & 78.9$^\dagger$ & 34.2 & 55.6 \\
    DeepSeek-V3.1  & 29.8$^\dagger$ & 35.4 & 83.7$^\dagger$ & 42.6$^\dagger$ & 65.0 \\
    \midrule

    \rowcolor[RGB]{229,229,252}\multicolumn{6}{c}{\emph{\textbf{Ours}}} \\
    \midrule
    \textbf{AgentFounder-30B}    & \textbf{31.5} & \textbf{{47.9}} & \textbf{89.6} & \textbf{43.9} & \textbf{75.3} \\
    \bottomrule
    \end{tabular}}
    \label{tab:scenario-targeted-web-search}
\end{table}

Table~\ref{tab:general-web-search}~and~\ref{tab:scenario-targeted-web-search} present a comprehensive performance comparison of our model AgentFounder-30B under the single-agent React paradigm against existing SOTA models. We have the following observations:

\textbf{Observations on General Web Search Benchmarks.} Overall, AgentFounder-30B outperforms all existing open-source deep research agents across four benchmarks and achieves comparable performance to DeepSeek-V3.1 on BrowseComp-zh. Moreover, it even surpasses commercial deep research agents on certain benchmarks. Specifically, on BrowseComp-en, AgentFounder-30B outperforms the best open-source model DeepSeek-V3.1 by 10.0\%, closely approaching the performance of OpenAI's closed-source o3 and Deep Research. This significant improvement demonstrates that AgentFounder-30B has effectively mastered sophisticated search strategies and reasoning capabilities.
Unfortunately, despite the similar question styles between BrowseComp-zh and BrowseComp-en, AgentFounder-30B's performance (43.3) on the Chinese version, while still surpassing strong open-source models such as GLM-4.5 (37.5), is comparable to DeepSeek-V3.1 (49.2) and remains behind OpenAI-o3 (58.1). Beyond the inherent distributional differences between BrowseComp-zh and BrowseComp-en evaluations, we attribute this performance gap to two potential reasons: the relatively limited proportion of Chinese data in our training corpus, and the possibility that the underlying search tool (Google Search) may exhibit suboptimal performance or bias in Chinese contexts.
On the remaining three benchmarks, AgentFounder-30B consistently outperforms all open-source deep research agents and even exceeds OpenAI-o3. Notably, AgentFounder-30B achieves the highest single-agent accuracy of 72.8\% on GAIA. Although this result is limited to GAIA's text subset, it nevertheless demonstrates that AgentFounder's capabilities extend beyond retrieval reasoning itself and can transfer to broader task categories, revealing its potential as a general-purpose agent in the future.

\textbf{Observations on Scenario-Targeted Web Search Benchmarks.} AgentFounder-30B demonstrates impressive performance across specialized evaluation tasks. On the highly challenging HLE benchmark, AgentFounder-30B becomes the first open-source model to surpass the 30-point threshold, achieving 31.5\%. This result significantly exceeds all reported closed-source deep research products, including Gemini-2.5-Pro Deep Research, Kimi-Researcher, and OpenAI Deep Research. 
Similarly, for academic capability assessment, AgentFounder-30B scores 75.3\% on Academic Browse, substantially outperforming all existing open-source models and demonstrating its value as an academic assistant. On the Frames benchmark, AgentFounder-30B substantially outperforms all open-source and closed-source models, demonstrating its superior capacity for multi-perspective reasoning and consistent information synthesis. In terms of robustness, AgentFounder-30B comprehensively outperforms open-source deep research agents on Seal-0, indicating strong resistance to information interference. Finally, on the DeepResearch Bench, AgentFounder-30B achieves 47.9\% on RACE Overall, surpassing both OpenAI Deep Research and all open-source deep research agents, confirming the comprehensiveness, readability, and depth of AgentFounder-30B's generated reports.







\subsection{Adaptability of the Agentic Base Model to Post-Training (RQ2)}
Agentic continual pre-training provides a pre-alignment base model that captures agentic behaviors prior to post-training. In this section, we investigate whether such a pre-alignment base model can effectively adapt to different paradigms of post-training.

\textbf{Experiment Design.} We adopt the AgentFounder-30B-Base as the pre-alignment agentic base model. We then conduct further agentic alignment training using three distinct SFT datasets as described in Section \ref{sec:exp_data}: {SFT-A}, {SFT-B}, and {SFT-C}. To ensure the robustness of results, we report {Pass@1} accuracy on four general web search benchmarks: {BrowseComp-en}, {BrowseComp-zh}, {GAIA}, and {Xbench-DeepSearch}.

\begin{table}[t]
\centering
\begin{tabular}{l l c c c c}
\toprule
\textbf{Base Model} & \textbf{SFT Data} & \textbf{BrowseComp-en} & \textbf{BrowseComp-zh} & \textbf{GAIA} & \textbf{HLE} \\
\midrule
Qwen3-30B-A3B-Base & SFT-A & 26.9 & 29.8 & 67.0 & 23.5 \\
AgentFounder-30B-Base   & SFT-A & \textbf{31.4} & \textbf{35.6} & \textbf{72.8} & \textbf{30.4} \\
\textit{Relative $\Delta$} & SFT-A & \textcolor{NavyBlue}{\textbf{+4.5}} & \textcolor{NavyBlue}{\textbf{+5.8}} & \textcolor{NavyBlue}{\textbf{+5.8}} & \textcolor{NavyBlue}{\textbf{+6.9}} \\
\midrule
Qwen3-30B-A3B-Base & SFT-B & 28.6 & 35.6 & 71.8 & 27.0\\
AgentFounder-30B-Base   & SFT-B & \textbf{39.9} & \textbf{43.3} & \textbf{72.8} & \textbf{31.5} \\
\textit{Relative $\Delta$} & SFT-B & \textcolor{NavyBlue}{\textbf{+11.3}} & \textcolor{NavyBlue}{\textbf{+7.7}} & \textcolor{NavyBlue}{\textbf{+1.0}} & \textcolor{NavyBlue}{\textbf{+4.5}} \\
\midrule
Qwen3-30B-A3B-Base & SFT-C & 24.5 & 36.7 & 68.9 & 27.9 \\
AgentFounder-30B-Base   & SFT-C & \textbf{38.8} & \textbf{44.3} & \textbf{71.8} & \textbf{28.9} \\
\textit{Relative $\Delta$} & SFT-C & \textcolor{NavyBlue}{\textbf{+14.3}} & \textcolor{NavyBlue}{\textbf{+7.6}} & \textcolor{NavyBlue}{\textbf{+2.9}} & \textcolor{NavyBlue}{\textbf{+1.0}} \\
\bottomrule
\end{tabular}
\caption{Adaptability Validation of the AgentFounder Base Model with Different Post-Training Data.}
\label{tab:posttraining-generalization}
\end{table}


\textbf{Observation Results.} Table~\ref{tab:posttraining-generalization} presents the results, from which we make the following three observations:

\textbf{(1) Agentic CPT demonstrates consistent and substantial improvements.} Models fine-tuned on AgentFounder-30B-Base consistently outperform their Qwen3-30B-A3B-Base counterparts across all configurations, validating the universal effectiveness of Agentic Continual Pre-Training. Specifically, models trained on SFT-A, SFT-B, and SFT-C datasets show average performance gains of 5.75\%, 6.13\%, and 6.45\% respectively when built upon AgentFounder-30B-Base. 

\textbf{(2) Post-training data remains crucial for unlocking base model capabilities.} Despite sharing the same AgentFounder-30B-Base foundation, models exhibit significant performance variations across different post-training datasets. For instance, on BrowseComp-zh, AgentFounder-30B with SFT-B outperforms variants with SFT-A and SFT-C by 8.5\% and 3.0\% respectively. This underscores the irreplaceable role of agentic post-training and highlights the ongoing challenge of fully realizing base model potential through optimal training strategies.

\textbf{(3) Information retrieval tasks benefit more from Agentic CPT than knowledge-intensive tasks.} Compared to HLE, the BrowseComp benchmarks show more pronounced improvements from AgentFounder-30B-Base. We hypothesize that knowledge-intensive tasks like HLE require not only successful information retrieval but also strong reasoning capabilities to correctly utilize the retrieved knowledge. This suggests that enhancing base models' knowledge comprehension abilities represents a promising future research direction.

\subsection{Ablation Studies}
\subsubsection{Impact of Training Strategies (RQ3)}
We examine whether our proposed two-stage training paradigm can bring performance improvements.

\textbf{Experiment Design.} We set the training tokens to 50B, using Qwen3-30B-A3B-Base as the initialization checkpoint and employing SFT-A data. We compare two configurations: AgentFounder Stage 1, which applies single-stage training on all data where some HAS data may be truncated due to length constraints, and AgentFounder Stage 1\&2, which implements our complete two-stage training approach that specifically incorporates synthesized long-context agent data in the second stage.

\begin{table}[t]
\centering
\setlength{\tabcolsep}{6pt}
\begin{tabular}{l
                cc cc
                cc cc
                cc cc}
\toprule
\multirow{2}{*}{\textbf{Strategy}} &
\multicolumn{4}{c}{\textbf{BrowseComp-en}} &
\multicolumn{4}{c}{\textbf{BrowseComp-zh}} &
\multicolumn{4}{c}{\textbf{GAIA}} \\
\cmidrule(lr){2-5} \cmidrule(lr){6-9} \cmidrule(lr){10-13}
& \multicolumn{2}{c}{Pass@1} & \multicolumn{2}{c}{Pass@3}
& \multicolumn{2}{c}{Pass@1} & \multicolumn{2}{c}{Pass@3}
& \multicolumn{2}{c}{Pass@1} & \multicolumn{2}{c}{Pass@3} \\
\midrule
AgentFounder Stage 1 Only 
& \multicolumn{2}{c}{31.4} & \multicolumn{2}{c}{49.9}
& \multicolumn{2}{c}{34.3} & \multicolumn{2}{c}{50.5}
& \multicolumn{2}{c}{69.9} & \multicolumn{2}{c}{81.6} \\
AgentFounder Stage 1 \& 2 
& \multicolumn{2}{c}{35.5} & \multicolumn{2}{c}{52.0}
& \multicolumn{2}{c}{37.2} & \multicolumn{2}{c}{58.5}
& \multicolumn{2}{c}{72.8} & \multicolumn{2}{c}{82.5} \\
\textit{Relative $\Delta$} 
& \multicolumn{2}{c}{\textcolor{NavyBlue}{\textbf{+4.1}}} & \multicolumn{2}{c}{\textcolor{NavyBlue}{\textbf{+2.1}}}
& \multicolumn{2}{c}{\textcolor{NavyBlue}{\textbf{+2.9}}} & \multicolumn{2}{c}{\textcolor{NavyBlue}{\textbf{+8.0}}}
& \multicolumn{2}{c}{\textcolor{NavyBlue}{\textbf{+2.9}}} & \multicolumn{2}{c}{\textcolor{NavyBlue}{\textbf{+0.9}}} \\
\bottomrule
\end{tabular}
\caption{Evaluation of the effectiveness of the two-stage AgentFounder training strategy.}
\label{tab:AgentFounder-train-strategy}
\end{table}



\textbf{Observation Results.} As shown in Table~\ref{tab:AgentFounder-train-strategy}, our analysis demonstrates that the two-stage training paradigm yields consistent performance improvements. Specifically, the AgentFounder Stage 1\&2 configuration achieves an average improvement of 3.3\% on Pass@1 and 3.7\% on Pass@3 across the three evaluation tasks compared to the single-stage baseline. These results substantiate the necessity of learning complete long-context agent data rather than truncated sequences. While resource constraints preclude evaluation of single-stage training with extended context lengths (e.g., 128K), such approaches would incur substantially higher computational costs.
\subsubsection{Impact of Data Types (RQ4)}
\begin{table}[t]
\centering
\begin{tabular}{l c c c c c c c}
\toprule
\multirow{2}{*}{\textbf{Data}} & \multirow{2}{*}{\textbf{Tokens}} & \multicolumn{2}{c}{\textbf{BrowseComp-en}} & \multicolumn{2}{c}{\textbf{BrowseComp-zh}} & \multicolumn{2}{c}{\textbf{GAIA}} \\
\cmidrule(lr){3-4} \cmidrule(lr){5-6} \cmidrule(lr){7-8}
 & & Pass@1 & Pass@3 & Pass@1 & Pass@3 & Pass@1 & Pass@3 \\
\midrule
Non CPT       & 0B    & 26.9 & 38.0 & 29.8 & 45.3 & 67.0 & 75.7 \\
FAS           & 50B  & 31.4 & 49.9 & 37.0 & 54.3 & 72.8 & 80.6\\
FAS+HAS       & 50B  & 31.4 & 50.1 & 40.1 & 54.7 & 69.9 & 82.5 \\
\textit{Relative $\Delta$} & /    & 0.0 & \textcolor{NavyBlue}{\textbf{+0.2}} & \textcolor{NavyBlue}{\textbf{+3.1}} & \textcolor{NavyBlue}{\textbf{+0.4}} & \textcolor{red}{\textbf{-2.9}} & \textcolor{NavyBlue}{\textbf{+1.9}} \\
\bottomrule
\end{tabular}
\caption{Effect of data type (HAS and FAS).}
\label{tab:data-sampling-strategy}
\end{table}

\textbf{Experiment Design.}  We investigate the effectiveness of the two data types introduced in this work for agentic CPT. We conduct single-stage training experiments using approximately 50B tokens of FAS data and FAS+HAS mixed data respectively, followed by post-training with SFT-A data. Performance is evaluated using Pass@1 and Pass@3 metrics on BrowseComp-en, BrowseComp-zh, and GAIA benchmarks.

\textbf{Observation Results.} Table~\ref{tab:data-sampling-strategy} presents our experimental results, revealing that both FAS and HAS data contribute meaningful performance improvements: \textbf{(1) FAS data demonstrates clear efficacy.} Training exclusively with FAS data yields substantial performance gains, particularly evident in Pass@3 metrics. Notably, on BrowseComp-zh, FAS-based continual pre-training achieves a 9.0\% improvement, establishing a higher performance ceiling for subsequent post-training phases. \textbf{(2) HAS provides complementary benefits.} The combination of FAS+HAS data consistently delivers positive gains across the evaluated benchmarks. While GAIA shows a modest 2.9\% decrease in Pass@1 performance, the corresponding 1.9\% improvement in Pass@3 suggests this variation falls within normal evaluation fluctuations rather than indicating systematic degradation.



\subsection{Scaling Law Exploration (RQ5)}

\subsubsection{Scaling with Model Size}
\textbf{Experiment Design}. We investigate how model scale influences agentic capabilities by evaluating models with 1B, 4B, and 30B parameters on deep research benchmarks. Additionally, we compare our models with two larger-scale baseline models (DeepSeek-V3.1 and Kimi-K2) to assess the efficiency of our agentic continual pre-training approach across different model scales.

\textbf{Observation Results}. Our experimental results reveal that model scale plays a crucial role in agentic performance, with our approach demonstrating superior scaling efficiency: 
\textbf{(1) Consistent scaling benefits}. Model size shows a strong positive correlation with agentic performance. Average accuracy increases from 20.4\% for the 1B model to 32.7\% for the 4B model and further to 48.9\% for the 30B model, indicating that larger models possess enhanced capacity for complex agentic behaviors such as effective tool use and multi-step reasoning. \textbf{(2) Superior scaling efficiency}. The AgentFounder-30B model achieves 48.9\% accuracy, exceeding the performance of two larger baseline models, DeepSeek-V3.1 (43.0\%) and Kimi-K2 (29.6\%), despite their greater size. This suggests that our agentic continual pre-training approach enables more effective utilization of model capacity for agentic tasks. These results demonstrate that combining agentic continual pre-training with appropriate model scaling provides a strong foundation for building high-performance deep research agents.  

\subsubsection{Scaling with Data Volume}
\begin{figure}[t]
    \centering
    \begin{minipage}{0.46\textwidth}
        \centering
        {\small (a) Model Scaling}
        \includegraphics[width=\linewidth]{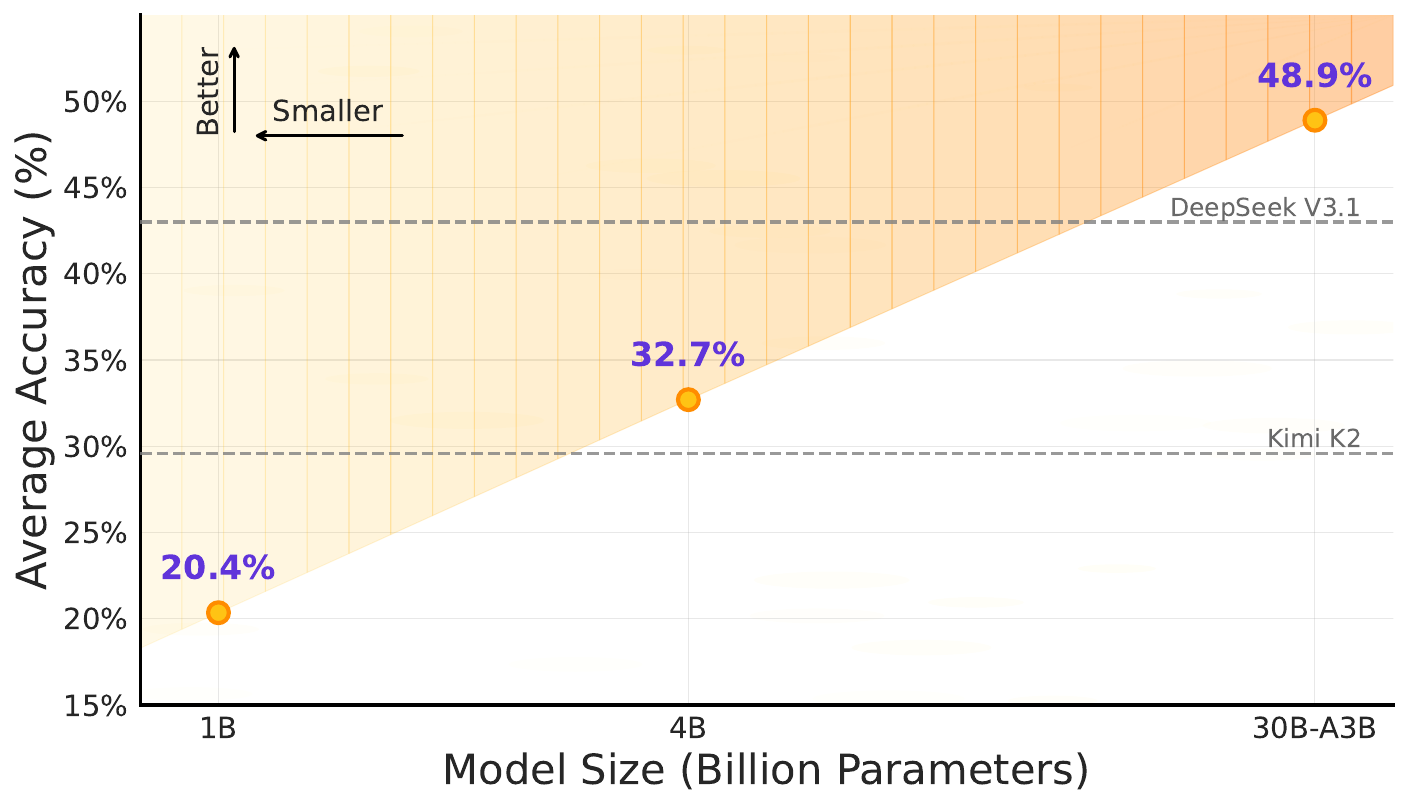}
    \end{minipage}
    \hfill
    \begin{minipage}{0.48\textwidth}
        \centering
        {\small (b) Data Scaling}
        \includegraphics[width=\linewidth]{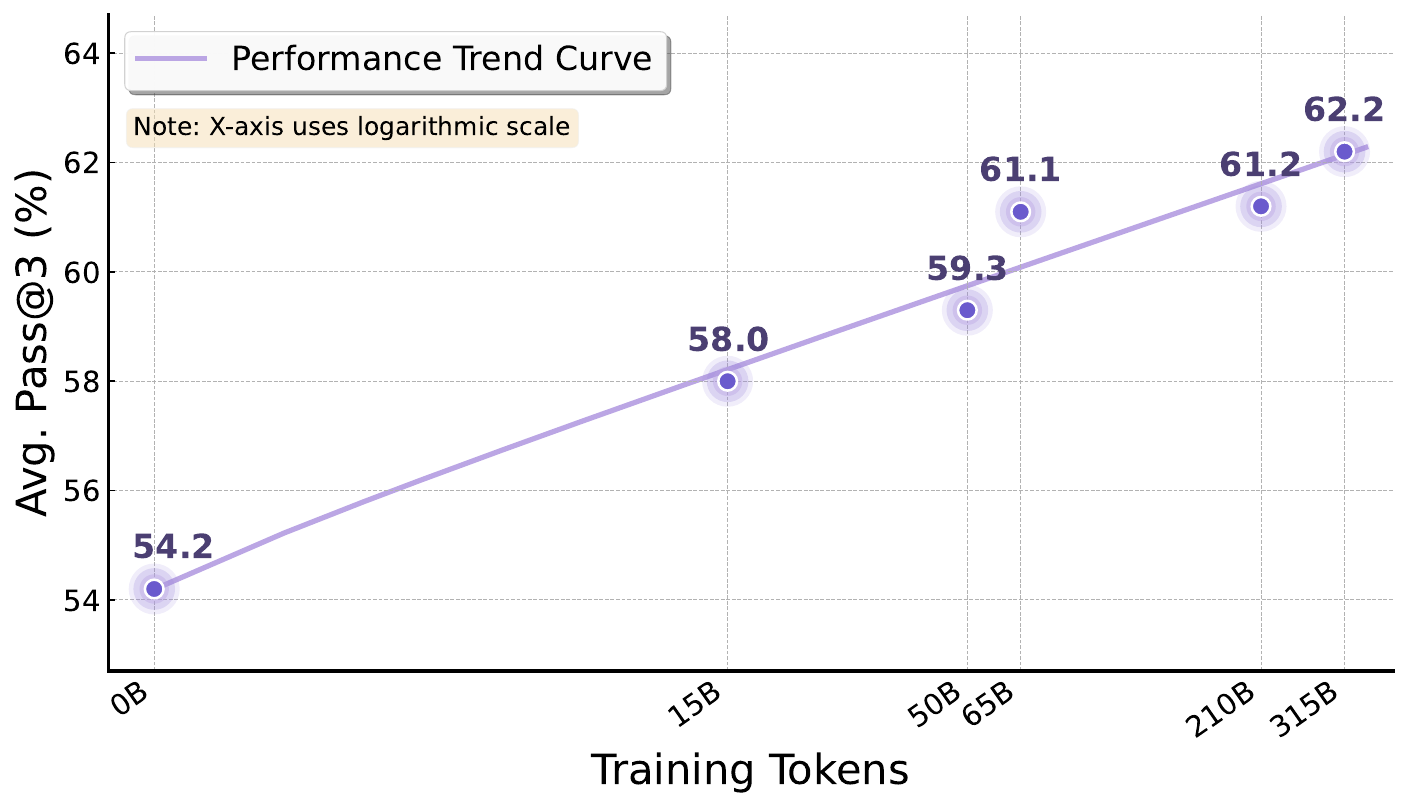}
    \end{minipage}
    \caption{Scaling Law Exploration for Agentic Capabilities. (a) Performance across model sizes ranging from 1B to 30B-A3B parameters. (b) Impact of training data volume from 0B to 315B samples on task performance. Both axes are in log scale.}
    \label{fig:scaling_size_data}
\end{figure}
We investigate the scaling properties of Agentic CPT and validate the effectiveness of our two-stage training strategy across varying data volumes.

\textbf{Experiment Design}. We train AgentFounder models with data volumes ranging from 0B to 315B tokens, using Qwen3-30B-A3B-Base as the initialization checkpoint. We implement our two-stage training paradigm, where Stage 2 incorporates 128K context window training at 65B and 315B token checkpoints. Performance is evaluated using the average Pass@3 metric across multiple agentic benchmarks to assess scaling behavior and training effectiveness.

\textbf{Observation Results}. As shown in the scaling curve, our analysis reveals three key findings:

\textbf{(1) Logarithmic scaling law holds for agentic capabilities.} The relationship between training tokens and performance exhibits logarithmic characteristics, with the most substantial improvements (3.8\%) occurring within the initial 15B tokens, demonstrating that agentic behaviors can be efficiently acquired through targeted pre-training. 

\textbf{(2) Two-stage CPT with extended context provides consistent improvements.} Stage 2 CPT with 128K context windows delivers notable gains at both 65B (+1.8\% over 50B) and 315B (+1.0\% over 210B), indicating that long-context training effectively enhances complex reasoning capabilities even when the base scaling curve approaches saturation. 

\textbf{(3) Sustained improvements at scale demonstrate robustness.} AgentFounder maintains consistent performance improvements throughout the entire 315B token budget, achieving a total gain of 8.0\% (from 54.2\% to 62.2\%), validating our training methodology's effectiveness in preventing premature convergence.





\subsection{Comprehensive Analyses}
\subsubsection{Training Process}
\begin{figure}[tb]
    \centering
    \includegraphics[width=0.6\textwidth]{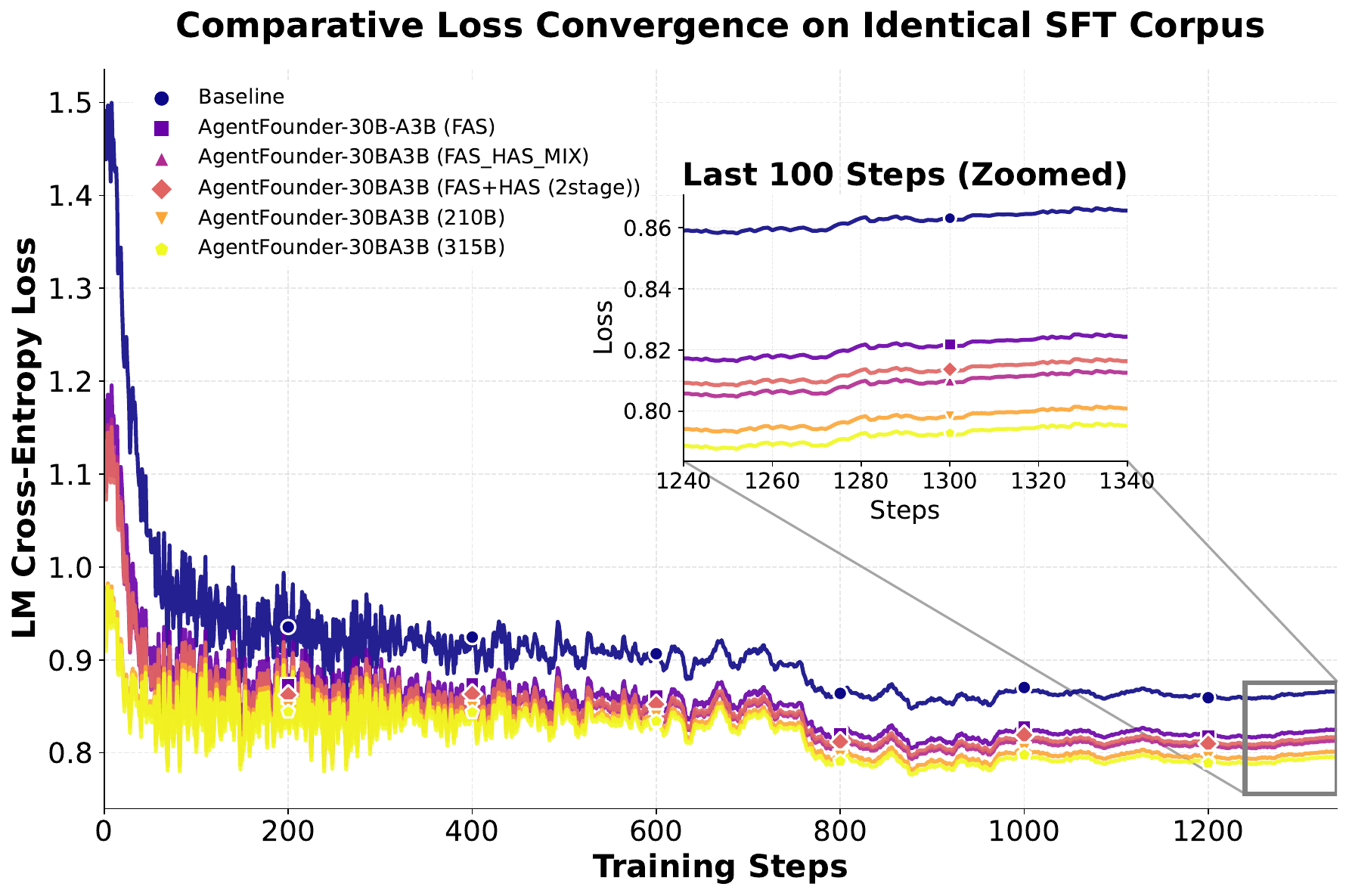}
    \caption{Training loss evolution showing superior convergence of AgentFounder models compared to baseline.}
    \label{fig:lm_loss_comparison}
\end{figure}
We validate whether agentic continual pre-training can alleviate the dual-burden problem by endowing models with foundational agentic capabilities before post-training.

\textbf{Experiment Design.} We conduct a comparative loss analysis between AgentFounder models and the baseline during SFT on identical downstream tasks. All models are trained for 1,340 steps using the same SFT-A data, with training efficiency measured through final loss, minimum achieved loss, and average loss over the last 100 training steps.

\textbf{Observation Results.} As shown in Figure~\ref{fig:lm_loss_comparison}, our analysis reveals that agentic CPT significantly enhances fine-tuning efficiency: \textbf{(1) AgentFounder substantially reduces SFT loss.} All AgentFounder variants achieve markedly lower loss values compared to the baseline across all metrics. While the baseline reaches a final loss of 0.8656, our best-performing AgentFounder-30B (315B) model achieves 0.7953, demonstrating that agentic CPT effectively endows models with foundational capabilities that facilitate adaptation to downstream agent tasks. \textbf{(2) Scaling CPT data yields monotonic improvements.} Loss values decrease progressively as CPT data volume increases from FAS-only to 315B tokens. The FAS+HAS mixture outperforms FAS alone, validating that reorganizing supervision signals from post-training into CPT format strengthens the model's agentic foundation.

\subsubsection{Tool Call Analysis}
\begin{figure}[h]
    \centering
    \begin{minipage}{0.48\textwidth}
        \centering
        {\small  WebWalker vs. GAIA}
        \includegraphics[width=\linewidth]{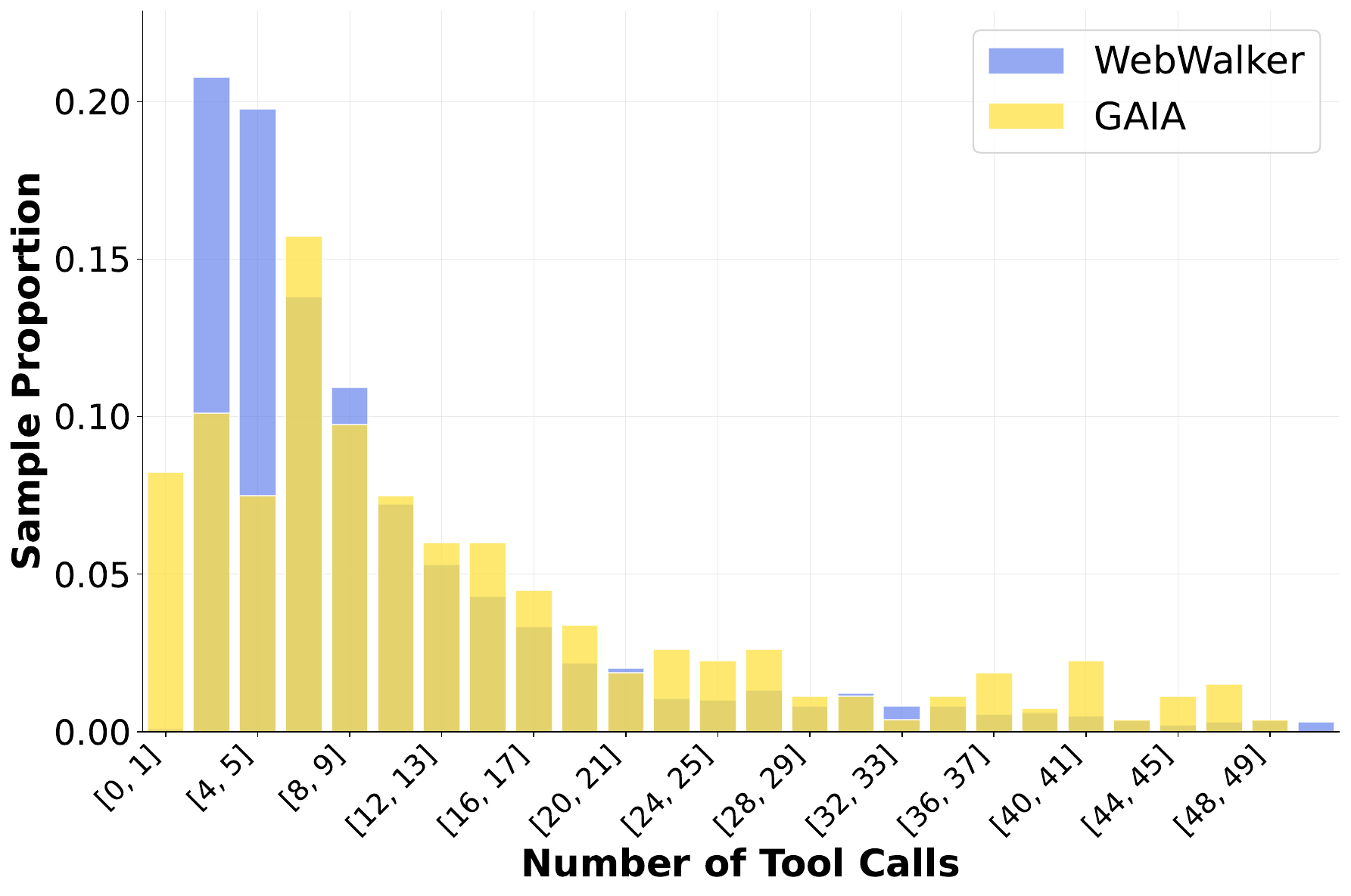}
    \end{minipage}
    \hfill
    \begin{minipage}{0.48\textwidth}
        \centering
        {\small HLE vs. BrowseComp(en)}
        \includegraphics[width=\linewidth]{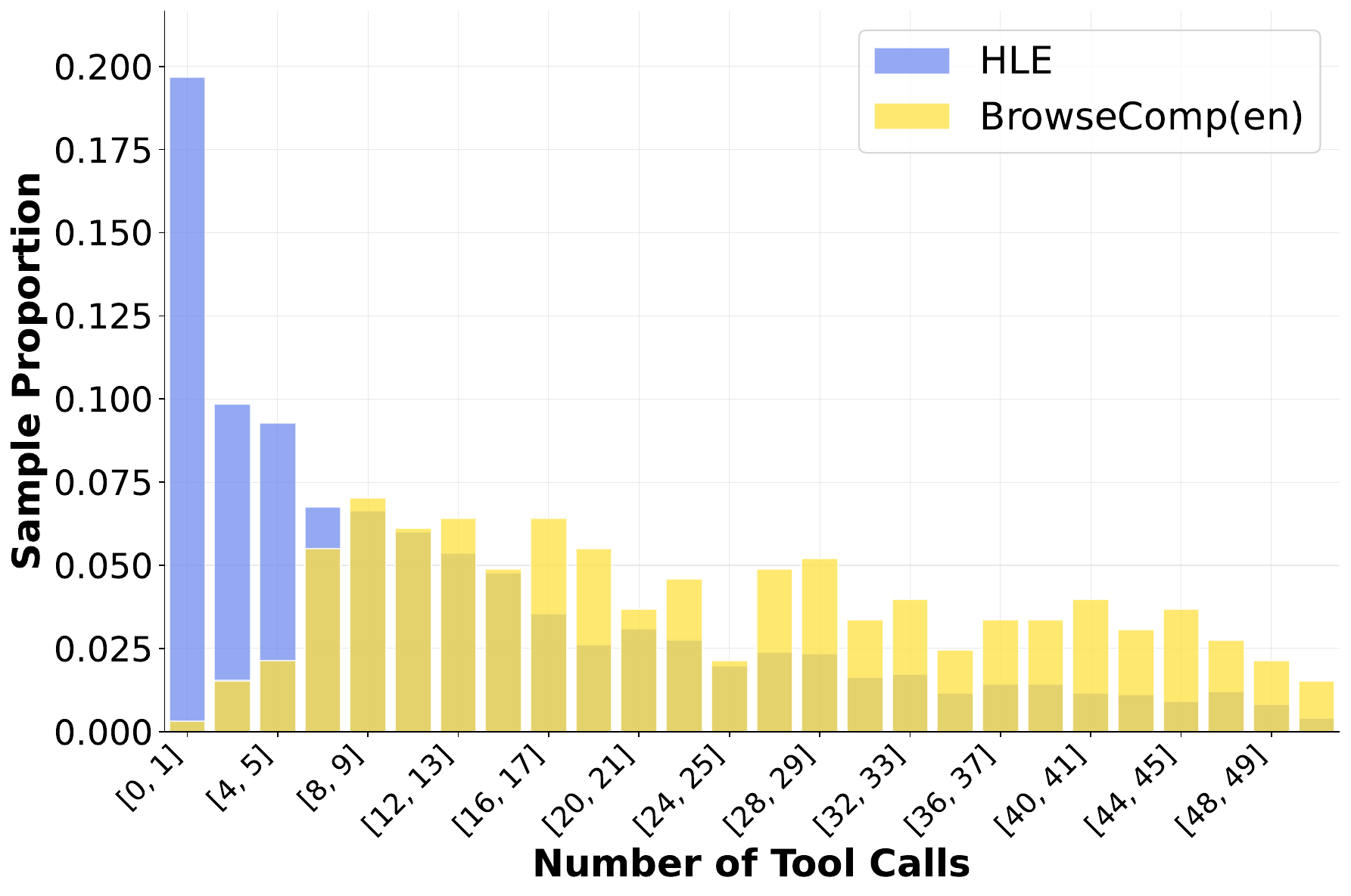}
    \end{minipage}
    \caption{Tool call distribution comparison.} 
    \label{fig:tool_call_distribution} 
\end{figure}
\textbf{Experiment Design.} We analyze AgentFounder's tool invocation distributions across four representative benchmarks to understand how the model adapts its tool usage strategies to different task complexities. We examine tool call patterns on HLE, BrowseComp-en, WebWalker, and GAIA-text benchmarks, categorizing them based on their task characteristics and required exploration depth.

\textbf{Observation Results.} As illustrated in Figure~\ref{fig:tool_call_distribution}, AgentFounder demonstrates distinct adaptive behaviors across task types: \textbf{(1) Complex research tasks exhibit intensive tool usage.} BrowseComp-en and HLE show heavy-tailed distributions with high tool density. BrowseComp-en requires extensive invocations for persistent web browsing, while HLE's extended patterns reflect the need to combine complex reasoning with search-augmented inference. \textbf{(2) Structured tasks employ conservative tool usage.} WebWalker's distribution peaks sharply at low invocation counts for efficient text navigation, while GAIA-text exhibits a compact distribution suited for well-defined problems with clear solution paths. These contrasting patterns demonstrate AgentFounder's ability to calibrate tool usage based on task complexity—intensive exploration for open-ended research versus targeted invocations for structured problems.

\begin{table}[h]
\centering
\begin{tabular}{l c }
\toprule
Model & ACEBench \\
\midrule
Qwen3-30B-A3B & 67.2  \\
AgentFounder-30B & 70.0 \\
\bottomrule
\end{tabular}
\caption{A comparison of general tool-use ability on ACEBench on overall performance.}
\label{tab:general-tool-use}
\end{table}

\subsubsection{General Tool-use Abilities} 
Beyond the tool-use capabilities of Deep Research, we construct agentic CPT data to address more general tool-use scenarios. As shown in Table \ref{tab:general-tool-use}, we compare AgentFounder-30B with Qwen3-30B-A3B on ACEBench \citep{chen2025acebenchwinsmatchpoint}, a benchmark designed to evaluate the tool-use performance of LLMs in diverse, general settings. The results reveal that AgentFounder-30B surpasses Qwen3-30B-A3B in general tool-use ability, suggesting that the Agentic CPT framework can be effectively applied to a broader range of agentic scenarios.

\section{Related Work}
\subsection{Deep Research Agents}
Deep research agents refer to language models that can autonomously invoke tools and perform multi-step reasoning, particularly utilizing search, web access, code execution, and file parsing capabilities, to complete research-level tasks or solve complex problems and generate trustworthy reports. OpenAI pioneered this field with their closed-source deep research agent, capable of searching and analyzing hundreds of web pages to generate comprehensive reports with citations within minutes \citep{openai2025deepresearch}. This breakthrough has catalyzed widespread industry adoption, with Grok \citep{grok3}, Perplexity \citep{perplexity2025deep}, Google Gemini \citep{google2025gemini_deep_research}, and Kimi \citep{moonshot2025kimiresearcher} subsequently releasing their own deep research products.

\textbf{Open-source deep research agents}. Open-source deep research agents have witnessed remarkable growth, achieving impressive results on deep research benchmarks including BrowseComp-en \citep{bc_en}, BrowseComp-zh \citep{bc_zh} Xbench-DeepSearch \citep{xbench}, and GAIA \citep{mialon2023gaia}. These works include early contributions such as Search-o1 \citep{li2025search}, R1-Searcher \citep{song2025r1}, EvolveSearch \citep{evolvesearch}, MaskSearch \citep{wu2025masksearch} and WebThinker \citep{li2025webthinker}, as well as later web agents that tackle challenging problems, such as MiroThinker \citep{2025mirothinker}. One common research focus across these works is how to construct more challenging problems and their corresponding trajectories for training agent models.
Accordingly, various innovative approaches have emerged for generating complex training data. WebDancer \citep{wu2025webdancer} and ASearcher \citep{gao2025beyond} take an iterative approach, incrementally adding new information to increase problem complexity. WebSailor \citep{li2025websailor} proposes SailorFog, a knowledge graph-based data synthesis method that starts with obscure Wikipedia entities and progressively builds entity relationship graphs through search and web reading to create highly ambiguous questions. WebShaper \citep{tao2025webshaper} adopts a more systematic methodology, formally modeling information-seeking tasks as set operations and introducing knowledge projection concepts to provide a principled method for constructing difficult problems. AFM \citep{afm} synthesizes high-quality agentic trajectory data by constructing a Chain-of-Agents pipeline. DeepDive \citep{deepdive} constructs high-difficulty QA pairs by randomly sampling nodes and their attributes within open knowledge graphs. Deepdiver \citep{deepdiver2} utilizes cross-page question generation and riddle creation to construct complex questions and assigns difficulty levels. WebExplorer \citep{liu2025webexplorerexploreevolvetraining} employs autonomous model exploration to build information networks, using iterative query evolution strategies to generate ambiguous and implicit questions.
While these methods successfully construct challenging post-training data that enhances agent capabilities for difficult problems, they predominantly follow supervised SFT or RL paradigms, overlooking the potential for training agentic capabilities during the continual pre-training phase. 

Recent open-source general models, including Kimi-K2 \citep{team2025kimi}, GLM-4.5 \citep{zeng2025glm}, and DeepSeek-V3.1 \citep{deepseekai2025deepseekv3_1}, have begun emphasizing enhanced agentic capabilities, yet the systematic exploration of continual pre-training for agent development remains limited.

\textbf{Multi Deep Research Agents and Multi-modal Deep Research Agents.} Beyond training methodologies, innovative inference paradigms have also emerged. Tencent's contribution extends beyond their trained CK-Pro-8B model to include the open-sourced Cognitive Kernel-Pro multi-agent framework \citep{fang2025cognitive}. Tencent Youtu has developed a multi-agent framework utilizing DeepSeek-V3.1, achieving 71.47\% accuracy on the WebWalkerQA benchmark \citep{webwalker}. Additional notable contributions include SkyworkAI's DeepResearchAgent \citep{AgentOrchestra} and ByteDance's deer-flow \citep{deerflow}, which further explore the potential of agent models. In the multimodal domain, Alibaba Tongyi Lab has open-sourced WebWatcher \citep{geng2025webwatcher}, the first multimodal deep research agent.

%




\subsection{Continual Pre-training}
While large language models are initially pretrained on massive general corpora, research shows that CPT can significantly enhance model performance by continuing unsupervised training \citep{gupta2023continualpretraininglargelanguage, lin2025rho1tokensneed, jin2022lifelongpretrainingcontinuallyadapting}.
\citet{ke2023continual} proposes continual domain-adaptive pre-training, which not only overcomes catastrophic
forgetting, but also achieves knowledge transfer to improve end-task performances. Though their experiments focus on million-scale language models on Natural Language Understanding tasks, the work gives insightful guidance for scaling CPT strategies to larger models and broader task families. 
Following that, \citet{yıldız2025investigatingcontinualpretraininglarge} investigates continual domain-adaptive pre-training in larger (billion-level) language models, aiming to enable models to assimilate new domain knowledge while preserving previously acquired information. Their experimental findings reveal that continual pre-training consistently enhances performance for models under 1.5B parameters and outperforms standard domain adaptation. 
\citet{parmar2024reusedontretrainrecipe} detail a set of guidelines about how to design efficacious data distributions and learning rate schedules for continued pre-training of language models. Upon applying the findings within a continued pre-training run on top of a
well-trained 15B parameter model, they find improvements compared to the baseline that is trained on the pre-training set.
These findings highlight CPT as a promising approach for expanding LLMs' agent capabilities.

However, regarding CPT and nowadays agents development, existing work focuses primarily either on CPT on tasks without tool calling or on post-training (such as SFT and RL) for agent development. Integrating agentic capabilities directly via the continual pre-training phase remains largely unexplored. This motivates our exploration of \textbf{Agentic CPT} as a novel paradigm that embeds agentic reasoning and tool-use capabilities at the foundational pre-training level.


%





\section{Conclusion}
In this work, we redefine the training pipeline for agentic alignment in deep research agents by introducing agentic continual pre-training (Agentic CPT) beyond traditional post-training for the first time. Furthermore, we present a systematic agentic CPT method comprising scalable agentic data synthesis and an adapted two-stage training strategy. Specifically for agentic data synthesis, we propose first-order action synthesis without additional commercial API calls, including planning action synthesis and reasoning action synthesis to enhance the model's planning and logical reasoning capabilities. Additionally, we provide a higher-order action synthesis method that remodels trajectories as multi-step decision-making problems, leveraging step-level expansion to thoroughly explore solution paths. Built on large-scale offline synthesized HAS and FAS data, we develop a powerful agentic model, AgentFounder-30B, which surpasses current closed-source models across 10 benchmarks, establishing a new state-of-the-art.


\vfill

\newpage

\clearpage
\appendix
\section{Experimental Setup Details}
\subsection{Tools}

\label{apx:tool}
In this work, we utilize five different tools for our AgentFounder model, namely Search, Visit, Python Interpreter, Google Scholar, and File Parser:
\begin{itemize}

    \item \textbf{Search} leverages the Google search engine for large-scale information retrieval. The tool accepts a list of one or more search queries to be executed concurrently. For each query, it returns the top-10 ranked results, with each result comprising a title, a descriptive snippet, and its corresponding URL.

    \item \textbf{Visit} is designed for targeted information extraction from web pages. The tool takes as input a set of web pages, where each page is paired with a dedicated information-seeking goal. The process begins by employing Jina~\citep{jina} to retrieve the full content of a given web page. Subsequently, a summary model processes this content to extract only the information pertinent to that page's specific goal. 


    \item \textbf{Python Interpreter} is used to execute Python code within a sandboxed environment. The input is a string of Python code, which must be enclosed within <code> tags for proper execution. The tool runs the provided code and captures its standard output; therefore, any results or values intended to be seen must be explicitly passed to the print() function. This capability enables dynamic computation, data manipulation, and the use of various Python libraries in a secure and isolated manner.

    \item \textbf{Google Scholar} is used to retrieve information from academic publications. The input consists of a list of one or more search queries, allowing for multiple, distinct searches within a single tool call. The tool leverages the Google Scholar search engine to execute each query and gather relevant scholarly literature, such as articles, papers, and citations.

    \item \textbf{File Parser} answers user queries by analyzing a mix of documents, web pages, and multimedia files (e.g., PDF, DOCX, MP4) from local or URL sources. It works in two steps: first, it converts all input into plain text, transcribing audio/video content when necessary. Second, a summary model reads this unified text to generate a direct answer to the user's question. 
    

\end{itemize}


\section{Additional Experimental Results}

\subsection{Quality Analysis of FAS data on Planning Action} \label{sec:fas_rej}
\begin{figure}[t]
    \centering
    \includegraphics[width=0.99\textwidth]{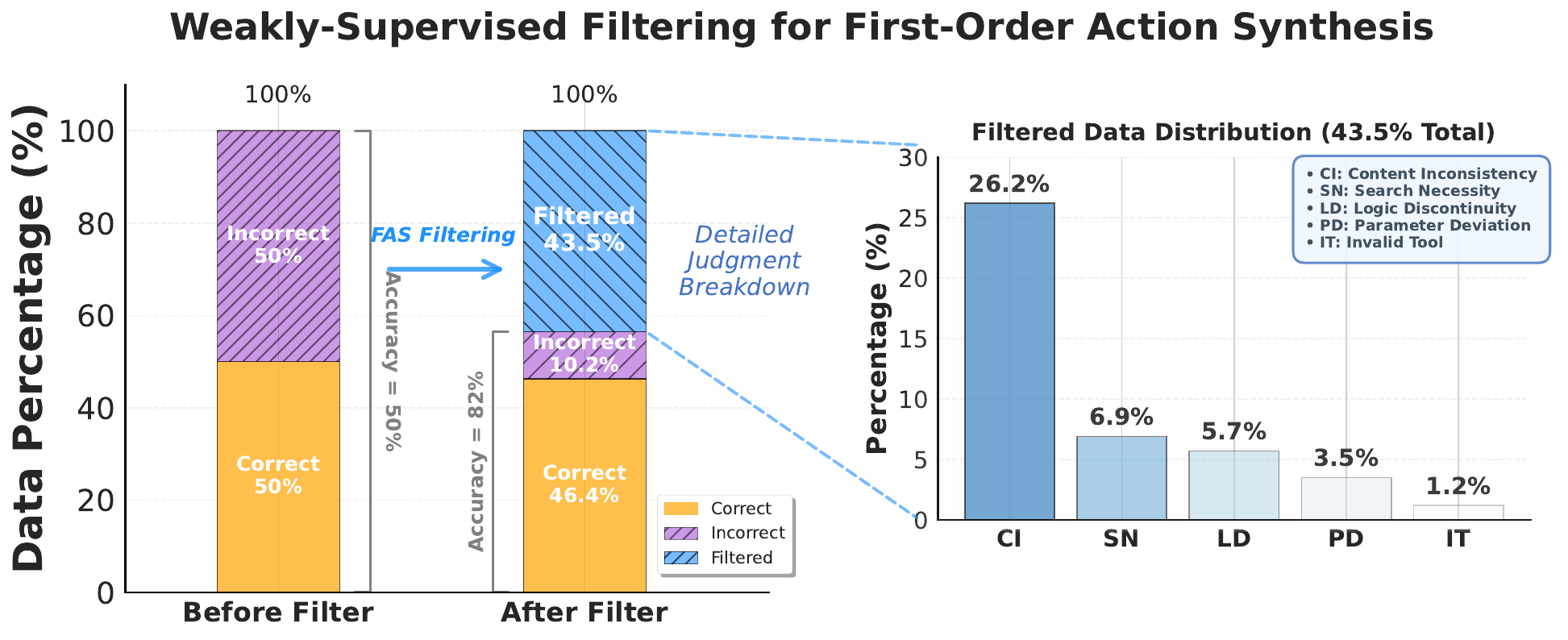}
    \caption{Filtering performance and representative low-quality outputs for weakly supervised filtering in first-order action synthesis.}
    \label{fig:fas_filter}
\end{figure}

\textbf{Experiment Design.} We evaluate the quality distribution and filtering effectiveness of FAS planning action data. Starting with FAS synthesized trajectories, we apply a prompt-based weak supervision filter that analyzes the original question, generated trajectory, and metadata to produce binary accept/reject decisions with detailed rejection reasons. We measure the impact on data quality by comparing pre- and post-filtering accuracy rates and analyzing error type distributions.

\textbf{Observation Results.} As shown in our analysis, the filtering mechanism significantly enhances FAS planning action data quality: \textbf{(1) Filtering achieves substantial quality improvement.} While initial FAS generation yields balanced correct/incorrect trajectories (50\%/50\%), our filter removes 43.5\% of problematic samples, increasing retained trajectory accuracy from 50\% to 82\%. This confirms FAS effectively generates diverse planning actions requiring quality control for high-fidelity selection. \textbf{(2) Semantic errors dominate rejection patterns}. Content Inconsistency accounts for 26.2\% of rejections, followed by Search Necessity (6.9\%) and Logic Discontinuity (5.7\%). The concentration in semantic rather than syntactic errors (Invalid Tool: 1.2\%) indicates FAS maintains structural validity while requiring refinement in semantic alignment. \textbf{(3) Quality improvement justifies volume reduction.} Though absolute correct data proportion slightly decreases (50\% → 46.3\%), the 82\% accuracy among retained samples validates prioritizing precision over recall for effective agentic CPT.

\begin{figure}[t]
    \centering
    \includegraphics[width=0.6\textwidth]{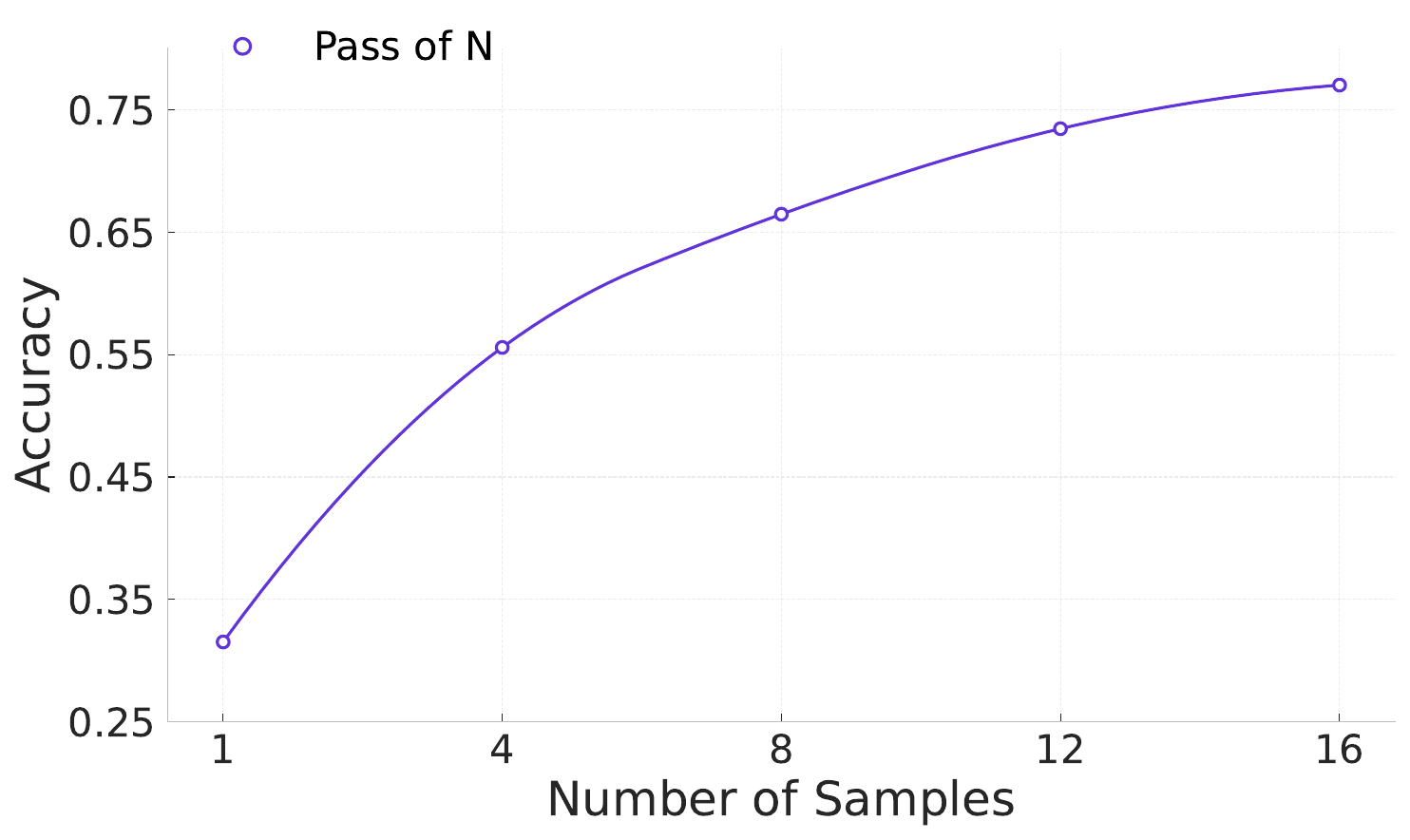}
    \caption{{Pass@N Scaling on BrowseComp-en.}}
    \label{fig:browsecomp_en_passn}
\end{figure}
\subsection{Sampling Strategies and Performance Scaling}

We examine whether HAS enhances behavioral diversity and enables effective scaling through sampling.

\textbf{Experiment Design.} We conduct Pass@n evaluation on BrowseComp-en with varying sampling sizes (n=1 to 18) using increased temperature to enable multiple trajectory generation. This experimental setup allows us to assess whether our HAS methodology, which exposes models to $(N+1) \times K$ potential actions per trajectory during training, successfully preserves solution diversity in the learned model.

\textbf{Observation Results.} As shown in Figure~\ref{fig:browsecomp_en_passn}, AgentFounder demonstrates strong scaling characteristics with progressive performance improvements: from 31.5\% Pass@1 to 75.8\% Pass@16, yielding a substantial +44.3 percentage point gain. The gradual saturation between Pass@16 and Pass@18 (only 1.16\% improvement) indicates a healthy balance between consistency and diversity. 

\subsection{Performance of GAIA on Different Levels}

\begin{figure}[tb]
    \centering
    \centering
        {\small  GAIA Pass@1 \& Pass@3 Analysis by Difficulty Level}
    \includegraphics[width=0.65\textwidth]{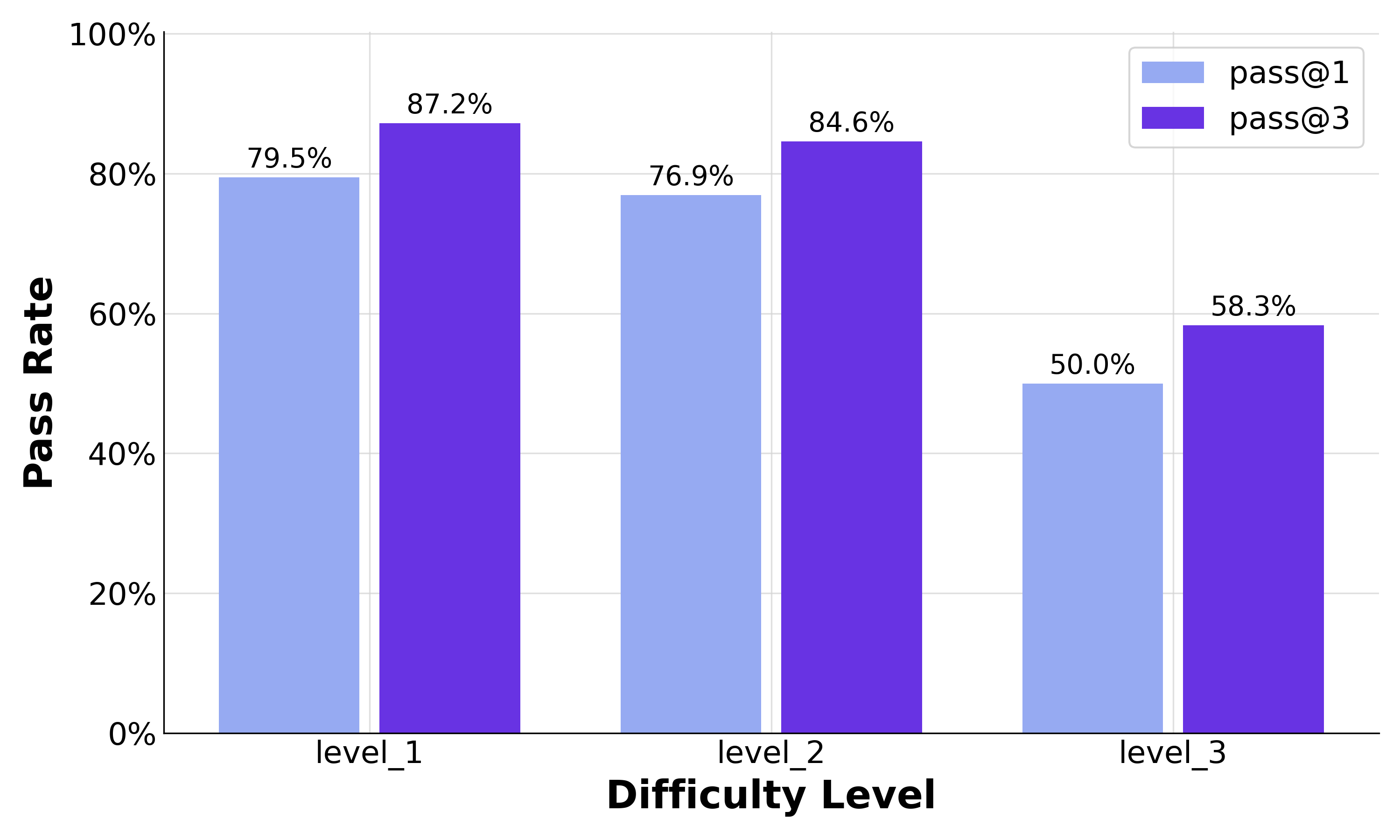}
    \caption{Pass rate on different levels of the GAIA dataset.}
    \label{fig:gaia_level123}
\end{figure}

We show the Pass Rate of the GAIA dataset w.r.t. different levels in Figure \ref{fig:gaia_level123}. Specifically, the model achieves its highest performance on level 1 tasks, with a 79.5\% Pass@1 rate and an 87.2\% Pass@3 rate. A more substantial performance degradation is evident at level 3, where the Pass@1 rate drops to 50.0\% and the Pass@3 rate falls to 58.3\%. This trend indicates that the model's efficacy is significantly impacted by the complexity of the tasks.

\begin{figure}[h]
    \centering
    \includegraphics[width=0.99\textwidth]{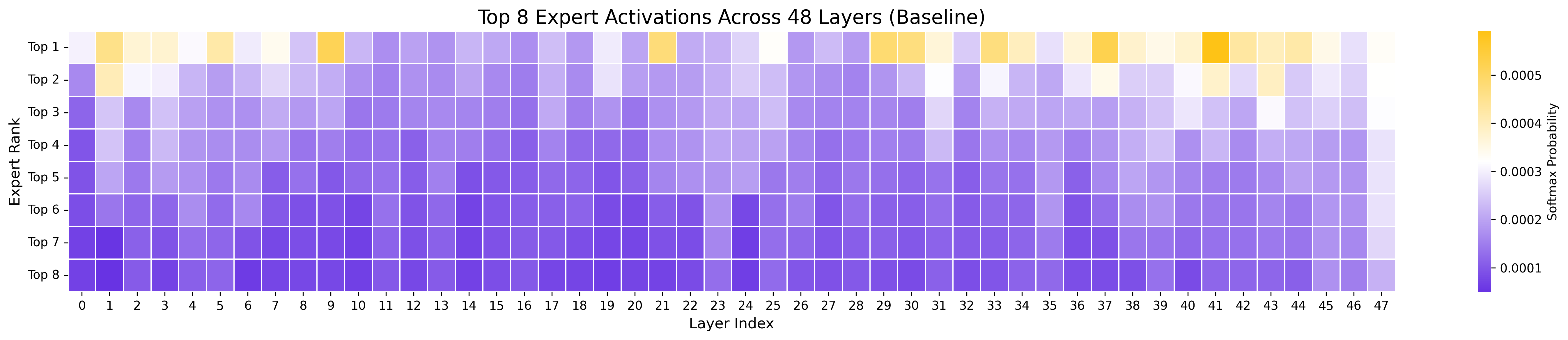}
    \includegraphics[width=0.99\textwidth]{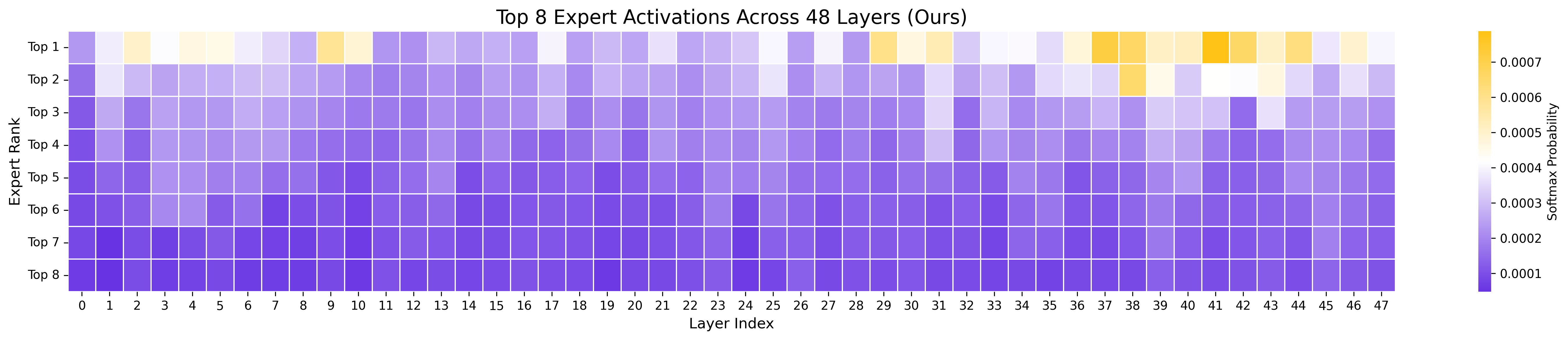}
    \includegraphics[width=0.99\textwidth]{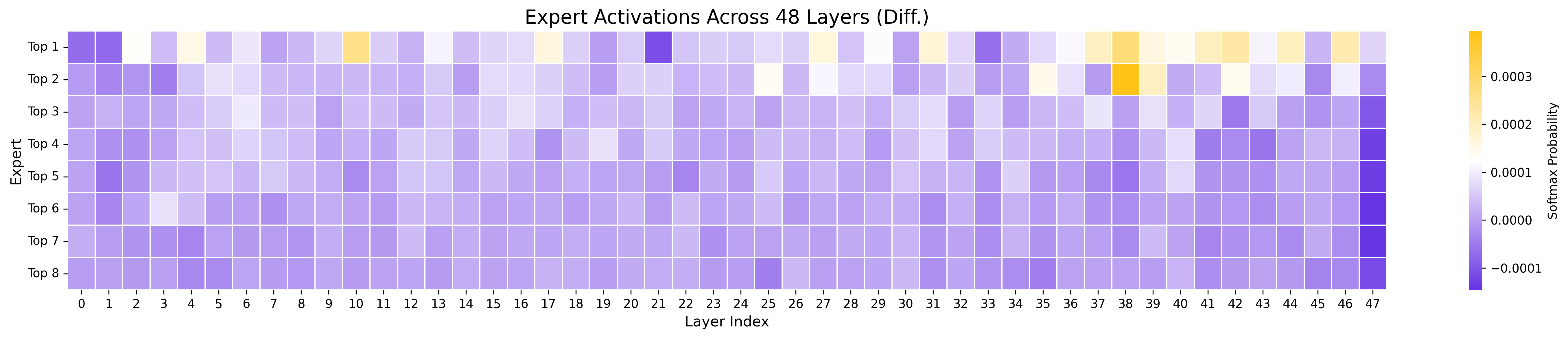}
    \caption{MoE activations illustration on BrowseComp-zh dataset. Top-Middle-Bottom: Baseline-Our model-Difference.}
    \label{fig:moe_act}
\end{figure}

\subsection{MoE Activations}

We collect the model's router logits for the questions' last token and display the top 8 ones after softmax. The scores are sorted and averaged over all samples in BrowseComp-zh.
The top one shows the activations of the baseline model without applying CPT. The middle one shows the result of our model AgentFounder-30B-A3B. The bottom one shows the difference after calculating ours minus the baseline. It can be observed that, after undergoing CPT, the expert distribution within the MoE architecture becomes more balanced in the final layers of the model, rather than being concentrated. This potentially offers the benefit of enabling more diversified utilization of multiple experts, which empirically leads to greater training stability during the post-training phase. It also helps mitigate the risk of overfitting by preventing the phenomenon of "dead experts"—experts that cease to be activated and thus fail to contribute meaningfully to the model’s performance.

\subsection{Accuracy vs. Tool Call Turns}

\begin{figure}[tb]
    \centering
    \small{Accuracy Distribution by Tool Call Turns (bin size = 8)}
    \includegraphics[width=0.9\linewidth]{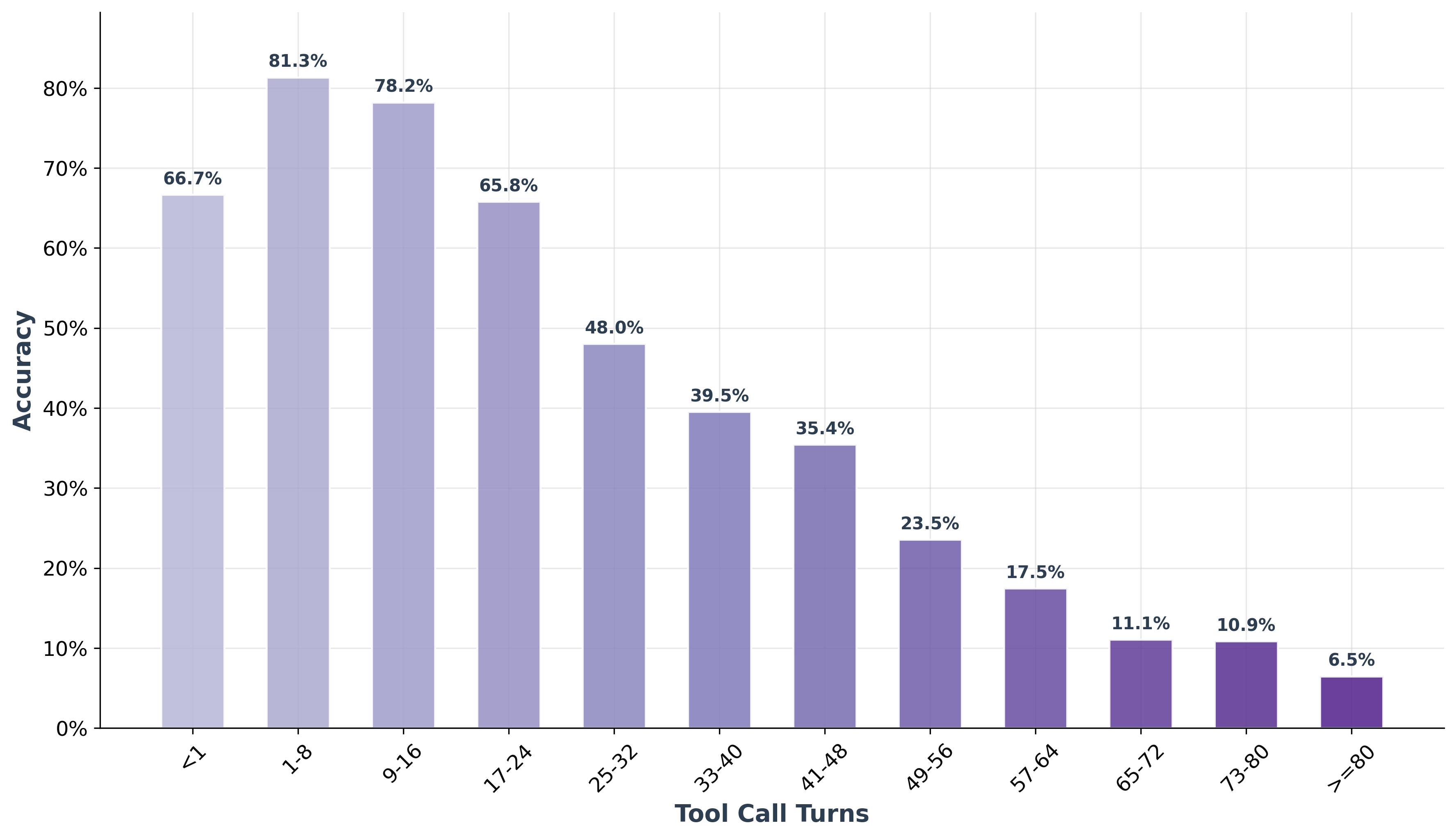}
    \caption{
        Accuracy distribution across tool call turns for \textit{BrowseComp-en}, \textit{BrowseComp-zh}, \textit{GAIA}, and \textit{Xbench-DeepResearch}, where darker colors indicate the number of tools used. The numeric value of each bin is annotated within its corresponding bar.
    }
    \label{fig:acc_tool_calls}
\end{figure}

To investigate the relationship between tool call frequency and final task success rate, we analyze rollout trajectories generated by our model on \textit{BrowseComp-en}~\citep{bc_en}, \textit{BrowseComp-zh}~\citep{bc_zh}, \textit{GAIA}\footnote{We use the text-only subset consisting of 103 questions.}~\citep{mialon2023gaia}, and the DeepResearch subset of \textit{Xbench}~\citep{xbench}. We measure the trajectories across three rollouts for stability.
For each trajectory, we record the total number of tool calls and whether the task was successfully completed.
We then aggregate the results into bins of size 8, as shown in Figure~\ref{fig:acc_tool_calls}, to visualize and analyze the correlation between tool usage and task outcomes. 

We observe three key trends: 
(1) Tasks with fewer tool call turns generally achieve higher final accuracy, indicating that the model succeeds more often in low-turn scenarios. This suggests improved planning capabilities, allowing more efficient task completion. 
(2) When no tool calls are made, accuracy drops noticeably (from 81.3\% to 66.7\%), and the corresponding sample size is very small, implying that the model benefits significantly from tool usage. 
(3) Even for more challenging cases, characterized by over 40 tool call turns, the model maintains a non-trivial success rate (average accuracy of 17.5\%), demonstrating its ability to perform in-depth research in complex environments.

\section{Case Study}
\begin{tcolorbox}[breakable,title=Case Trajectory]
\textcolor{black}{\textbf{Question:} At the biennial aerospace marketplace named after the city whose pyramid-fronted museum recorded high single-digit millions of visitors during a period of global athletic celebration, and where the year before a citywide nuisance led authorities to convene transit operators, which buyer placed a perfectly balanced commitment with firm orders equal to options? }\\
\colorbox{gray!70}{%
    \parbox{\dimexpr\textwidth-2\fboxsep}{%
    \centering
        \textbf{Step 1}%
    }%
}

\textbf{<think>} \\
...
So we need to identify:\\
\textcolor{purple1}{- A biennial aerospace marketplace named after a city.\\
- That city had a pyramid-fronted museum that recorded high single-digit millions of visitors during a period of global athletic celebration (maybe Olympic Games?).\\
- "High single-digit millions of visitors" probably means 7-9 million visitors, maybe 8 million. The museum has a pyramid front. Which museum has a pyramid? There is the Louvre Pyramid in Paris: the Louvre Museum has a glass pyramid entrance; it gets about 10 million visitors per year historically (pre-COVID). But high single-digit millions? Actually the Louvre had around 9.5 million visitors in 2019 (maybe earlier?). Also "global athletic celebration" could be Olympics. The 2016 Olympics were in Rio de Janeiro; but Rio's museum? The city might be Rio de Janeiro; it has the Museum of Tomorrow, but not pyramid. Another possibility: The city is Beijing (the National Museum?), the Bird's Nest stadium for Olympics? No.\\}

...Maybe they refer to the "Louvre Abu Dhabi"? It has a dome, not pyramid. The "Kunsthistorisches Museum"? Not.
There's the "Museum of the Iraqi Ministry"? Hmm.
Better to search. Let's search web for \textcolor{purple1}{"pyramid front museum high single digit millions visitors Olympic"}.\\
\textbf{</think>} \\
\textbf{<tool\_call>} \\
{'query': ['pyramid front museum high single digit millions visitors Olympics', 'pyramid front museum visitors 9 million', 'pyramid front museum visitor numbers Olympic Games', 'pyramid front museum recorded high single-digit millions visitors during Olympics']} \\
\textbf{</tool\_call>}\\
\textbf{<tool\_response>}\\
A Google search for 'pyramid front museum high single digit millions visitors Olympics' found 10 results:\\
Web Results \\
1. [Louvre](\url{https://en.wikipedia.org/wiki/Louvre})
It received 8.7 million visitors in 2024, 200,000 less than 2023... \\
... 

A Google search for 'pyramid front museum visitors 9 million' found 10 results:\\
Web Results \\
1. [Not just the Louvre: How Europe's museums face the future](\url{https://www.dw.com/en/not-just-the-louvre-how-europes-museums-are-facing-the-future/a-71471539})
If you haven't visited the Louvre, then you haven't... \\
...

% ...

\textbf{</tool\_response>}

\colorbox{gray!70}{%
    \parbox{\dimexpr\textwidth-2\fboxsep}{%
    \centering
        \textbf{Step 2}%
    }%
}
\textbf{<think>} \\
...\textcolor{purple1}{The museum recorded high single-digit millions of visitors during a period of global athletic celebration – likely referring to the Paris Olympics in 2024}
... 
This refers to some incident in Paris a year before something? "Citywide nuisance" possibly referencing noise or something that caused public disruption. Let's search for "citywide nuisance led authorities to convene transit operators Paris"\\
\textbf{</think>} \\
\textbf{<tool\_call>} \\
{'query': ['citywide nuisance led authorities to convene transit operators Paris year before', 'Paris citywide nuisance transit operators convened', 'Paris citywide nuisance transit operators meeting']}\\
\textbf{</tool\_call>}\\
\textbf{<tool\_response>}\\
A Google search for 'citywide nuisance led authorities to convene transit operators Paris year before' found 10 results:\\
Web Results \\
1. [Innovative solutions for sustainable cities]\\(\url{https://www.ecologie.gouv.fr/sites/default/files/documents/mobility\_gb.pdf})
Date published: 
From long time, France has built up a tightly-meshed mass transit system, upgraded and adapted to new forms of mobility: coverage of mass transit in the...\\
...

A Google search for 'Paris citywide nuisance transit operators convened' found 10 results:\\
Web Results \\
1. [Electric Rental Scooters to be Banned in Paris ...] \\(\url{https://www.nytimes.com/2023/04/03/world/europe/paris-electric-scooters-ban.html})
A referendum emphasized how many residents...\\
...

% ...

\textbf{</tool\_response>}\\
\colorbox{gray!70}{%
    \parbox{\dimexpr\textwidth-2\fboxsep}{%
    \centering
        \textbf{Step 3}%
    }%
}
\textbf{<think>} \\
The "citywide nuisance" could be referring to the electric scooters, described as a nuisance and 
...\textcolor{purple1}{due to the scooter issue. Let's search for article referencing "authorities convened transit operators" after "citywide nuisance"}
.\\
\textbf{</think>} \\
\textbf{<tool\_call>} \\
{'query': ['"convened transit operators" Paris "electric scooters"', '"convene" "transit operators" Paris "scooter"']} \\
\textbf{</tool\_call>}\\
\textbf{<tool\_response>}\\
A Google search for '"convened transit operators" Paris "electric scooters"' found 10 results:
\\
Web Results \\
1. [Rented e-scooters cleared from Paris streets on eve of ban](\url{https://www.theguardian.com/world/2023/aug/31/rented-e-scooters-cleared-from-paris-streets-on-eve-of-ban})Paris will become the first European capital to...\\
...

A Google search for '"convene" "transit operators" Paris "scooter"' found 10 results:
\\
Web Results \\
1. [U.S. National Blueprint for Transportation Decarbonization ...](\url{https://www.energy.gov/sites/default/files/2023-01/the-us-national-blueprint-for-transportation-decarbonization.pdf})
Mobile applications can more seamlessly integrate multiple...\\
...

\textbf{</tool\_response>}\\
\colorbox{gray!70}{%
    \parbox{\dimexpr\textwidth-2\fboxsep}{%
    \centering
        \textbf{Step ...}%
    }%
}
...\\
\colorbox{gray!70}{%
    \parbox{\dimexpr\textwidth-2\fboxsep}{%
    \centering
        \textbf{Step 25}%
    }%
}

\textbf{<think>} \\
...\textcolor{purple1}{That could be considered perfectly balanced commitment (firm orders equal to options) for Avelo Airlines. But question: "which buyer placed a perfectly balanced commitment with firm orders equal to options?" Let's read that article}.
 \\
\textbf{</think>} \\
\textbf{<tool\_call>} \\
{'url'['\url{https://leehamnews.com/2025/09/10/embraer-e2-finally-breaks-into-the-u-s-market-following-50-firm-orders-50-purchase-rights-from-ulcc-avelo-airlines/}'], 'goal': 'Extract details about buyer and number of firm orders vs options', 'url\_info': [{'idx': 3, 'query': '"firm orders" "options" "equal" "Paris Air Show"', 'url': '\url{https://leehamnews.com/2025/09/10/embraer-e2-finally-breaks-into-the-u-s-market-following-50-firm-orders-50-purchase-rights-from-ulcc-avelo-airlines/}', 'title': 'Embraer E2 Finally Breaks Into The U.S. Market Following ...', 'snippet': ' options for five more at the 2025 Paris Air Show, with deliveries beginning in 2028. SAS followed with its largest aircraft order in nearly ', 'hostlogo': None, 'date': '5days ago', 'source': None, 'id': 166}]} \\
\textbf{</tool\_call>}\\
\textbf{<tool\_response>}\\
The useful information in \url{https://leehamnews.com/2025/09/10/embraer-e2-finally-breaks-into-the-u-s-market-following-50-firm-orders-50-purchase-rights-from-ulcc-avelo-airlines/}


...Summary:
Avelo Airlines, an Ultra-Low-Cost Carrier, has placed a firm order for 50 Embraer E2-E195 aircraft and has 50 additional purchase rights. The aircraft are scheduled to enter service with Avelo in mid-2027, with deliveries continuing until 2032. \\
\textbf{</tool\_response>}\\
\colorbox{gray!70}{%
    \parbox{\dimexpr\textwidth-2\fboxsep}{%
    \centering
        \textbf{Step ...}%
    }%
}
...


% ...

\colorbox{gray!70}{%
    \parbox{\dimexpr\textwidth-2\fboxsep}{%
    \centering
        \textbf{Step 49}%
    }%
}
\textbf{<think>} \\
Not helpful. However we have enough evidence to answer.
Thus answer: Riyadh Air.
But perhaps the question expects "Riyadh Air" is the buyer. 
\textcolor{purple1}{Let's verify the official naming: "Riyadh Air" (the new Saudi airline)}.
The order was for 25 Airbus A350-1000s with options for an additional 25, which is perfectly balanced. So answer: Riyadh Air...
\\
\textbf{</think>} \\
\textbf{<tool\_call>} \\
{'query': ['"ASL Aviation" "firm orders" "options" "Paris Air Show"']} \\
\textbf{</tool\_call>}\\
\textbf{<tool\_response>}\\
A Google search for '"ASL Aviation" "firm orders" "options" "Paris Air Show"' found 10 results: \\
Web Results \\
1. [Fleet Renewal - ASL Aviation Holdings](\url{https://www.aslaviationholdings.com/fleet-renewal/})
At ASL Aviation  firm orders and 10 purchase rights, announced at the Paris Air Show in Le Bourget in June 2019...\\
...

\textbf{</tool\_response>}\\
\colorbox{gray!70}{%
    \parbox{\dimexpr\textwidth-2\fboxsep}{%
    \centering
        \textbf{Step 50}%
    }%
}
\textbf{<think>} \\
...
Let's examine context: "the year before a citywide nuisance led authorities to convene transit operators"...
Now, the event where the buyer placed a perfectly balanced commitment with firm orders equal to options might be at the same air show (Paris Air Show). In 2025, Riyadh Air placed a perfect balance of 25+25.

Alternatively, in 2019 (previous Paris Air Show), ASL Aviation placed 10+10...
\textcolor{purple1}{So maybe the question expects us to answer about Riyadh Air because it's the latest.However we need to consider that "the year before a citywide nuisance led authorities to convene transit operators" indicates that the nuisance event happened the previous year relative to the statement}. 
\textcolor{purple1}{Then ask: which buyer placed a perfectly balanced commitment at the biennial aerospace marketplace? That suggests the buyer placed that order at the Paris Air Show 2025 (recent). Therefore Riyadh Air is correct}
.

Thus answer: Riyadh Air.

We need to provide answer clearly: The buyer is Riyadh Air, which placed a firm order for 25 Airbus A350-1000 aircraft with purchase rights for an additional 25 (equal).

Thus final answer: Riyadh Air. \\
\textbf{</think>} \\
\textcolor{red}{\textbf{<answer>}} \\
The question references several interconnected events centered around Paris...:\\
1. \textbf{Biennial Aerospace Marketplace}: ...\textcolor{purple1}{Paris Air Show, officially known as} ... \textcolor{purple1}{attracting manufacturers, airlines, buyers, and aviation professionals from across the world}...

2. \textbf{Pyramid-Fronted Museum and High Single-Digit Visitor Numbers During Global Athletic Celebration}: ... \textcolor{purple1}{In 2024, during the period of the Paris Olympic Games—a major global athletic celebration—the Louvre recorded visitor numbers in the high single-digit millions (approximately 8.7 million visitors in 2023 and similar figures in subsequent months)}...

3. \textbf{Citywide Nuisance and Convening of Transit Operators}: In 2023, Paris experienced a citywide debate on electric scooters (e-scooters), \textcolor{purple1}{which many residents and officials labeled a “nuisance” due to safety concerns and congestion}...


\textbf{Recent Developments at the Paris Air Show}

... \textcolor{purple1}{Among the many deals announced, Riyadh Air stood out for placing a perfectly balanced commitment... This means the number of firm orders equals the number of options, making the commitment symmetrical and strategically flexible}.

Key details of Riyadh Air’s order:...

\textbf{Historical Context and Other Buyers
}
... \textcolor{purple1}{While this demonstrated a similarly balanced commitment, the query appears to focus on recent activity aligning with the context provided (i.e., after the citywide e-scooter nuisance debate and Olympic Games)}.\\

\resizebox{\linewidth}{!}{
\begin{tabular}{|l|c|l|c|c|}
\hline
\textbf{Buyer} & \textbf{Year Announced} & \textbf{Aircraft Model} & \textbf{Firm Orders} & \textbf{Options} \\ \hline
Riyadh Air & 2025 & Airbus A350-1000 & 25 & 25 \\ \hline
ASL Aviation Holdings DAC & 2019 & Boeing 737-800BCF & 10 & 10 \\ \hline
\end{tabular}
} 
\vspace{0.8em}






\textbf{Conclusion}

In summary, the buyer that placed a perfectly balanced commitment with firm orders equal to options at the recent Paris Air Show is Riyadh Air... \\
\textcolor{red}{\textbf{<answer>}} \\
\textcolor{black}{\checkmark}\\

\end{tcolorbox}

\clearpage
\bibliography{biblio}
\bibliographystyle{colm2024_conference}

\end{document}